\pdfoutput=1

\documentclass[11pt]{article}

\usepackage[final]{acl}

\usepackage{times}
\usepackage{latexsym}
\usepackage[T1]{fontenc}
\usepackage[utf8]{inputenc}

\usepackage{microtype}
\usepackage{inconsolata}
\usepackage{amsmath}
\usepackage{graphicx}
\usepackage{subfig}
\usepackage{multirow}
\usepackage{graphicx}
\usepackage{xcolor,colortbl}
\usepackage[many]{tcolorbox}
\usepackage{soul} % 导入 soul 包

\usepackage{ctable}
\usepackage{enumitem}
\usepackage[linesnumbered,lined,boxed,commentsnumbered,ruled,vlined]{algorithm2e}
\usepackage{booktabs}
\let\oldnl\nl
\newcommand{\nonl}{\renewcommand{\nl}{\let\nl\oldnl}}

\title{Beyond Demographics: Enhancing Cultural Value Survey Simulation \\with Multi-Stage Personality-Driven Cognitive Reasoning}

\newcommand{\cuhksz}{$^1$}
\newcommand{\hulab}{$^2$}
\newcommand{\hkust}{$^3$}
\newcommand{\tubingen}{$^4$}
\newcommand{\ku}{$^5$}

\author{Haijiang Liu\cuhksz \hulab \hkust \thanks{Work done during his visit at The Hong Kong University of Science and Technology (Guangzhou).}, 
Qiyuan Li\cuhksz \hulab, 
Chao Gao\cuhksz \hulab, 
Yong Cao\tubingen, 
\\\textbf{Xiangyu Xu}\hkust, 
\textbf{Xun Wu}\hkust, 
\textbf{Daniel Hershcovich}\ku, 
\textbf{Jinguang Gu}\cuhksz \hulab \thanks{Corresponding Author: Jinguang Gu. (E-mail:
simon@ontoweb.wust.edu.cn).} \\
{\small {\cuhksz} School of Computer Science and Technology, Wuhan University of Science and Technology, Wuhan 430065, China} \\[-4pt]
{\small {\hulab} Hubei Province Key Laboratory of Intelligent Information Processing and Real-time Industrial System,}\\[-4pt] {\small Wuhan University of Science and Technology, Wuhan 430065, China} \\[-4pt]
{\small {\hkust}Innovation, Policy and Entrepreneurship Thrust,} \\[-4pt] {\small The Hong Kong University of Science and Technology (Guangzhou), Guangzhou, 511453, China} \\[-4pt]
{\small {\tubingen}Tübingen AI Center, University of Tübingen, Tübingen, 72074, Germany} \\[-4pt]
{\small {\ku}Department of Computer Science, University of Copenhagen, 2100 Copenhagen, Denmark}\\ [-4pt]
{\small \{\texttt{alecliu}, \texttt{simon}\} \texttt{@ontoweb.wust.edu.cn}}
}

\begin{document}
\definecolor{lightorange}{HTML}{fad7ac}
\definecolor{lightyellow}{HTML}{fff2cc}
\definecolor{lightpurple}{HTML}{e1d5e7}
\definecolor{lightblue}{HTML}{d1e5ec}
\definecolor{lightgreen}{HTML}{cadfc8}

\definecolor{main}{HTML}{5989cf}
\definecolor{sub}{HTML}{cde4ff}
\newtcolorbox{boxH}{
    colback = sub, 
    colframe = main, 
    boxrule = 0pt, 
    leftrule = 6pt 
}

\maketitle

\begin{abstract}
Introducing \textbf{MARK}, the \textbf{M}ulti-st\textbf{A}ge \textbf{R}easoning framewor\textbf{K} for cultural value survey response simulation, designed to enhance the accuracy, steerability, and interpretability of large language models in this task. The system is inspired by the type dynamics theory in the MBTI psychological framework for personality research. It effectively predicts and utilizes human demographic information for simulation: life-situational stress analysis, group-level personality prediction, and self-weighted cognitive imitation.
Experiments on the World Values Survey show that MARK outperforms existing baselines by 10\% accuracy and reduces the divergence between model predictions and human preferences. This highlights the potential of our framework to improve zero-shot personalization and help social scientists interpret model predictions.\footnote{\url{https://github.com/alexc-l/MARK-Multi-stAge-Reasoning-frameworK}} 
% of AI-driven simulations in social survey contexts.
\end{abstract}

\section{Introduction}
Social surveys are essential for generating insights and supporting decision-making processes in research, yet traditional human-centric experiments often incur significant costs. Therefore, recent studies leverage AI technologies with human-behaviour simulating abilities to create more accurate and interpretable simulations as a valuable alternative \citep{horton2023large,argyle2023out}.

\begin{figure}[t]
    \centering
        \includegraphics[width=\linewidth]{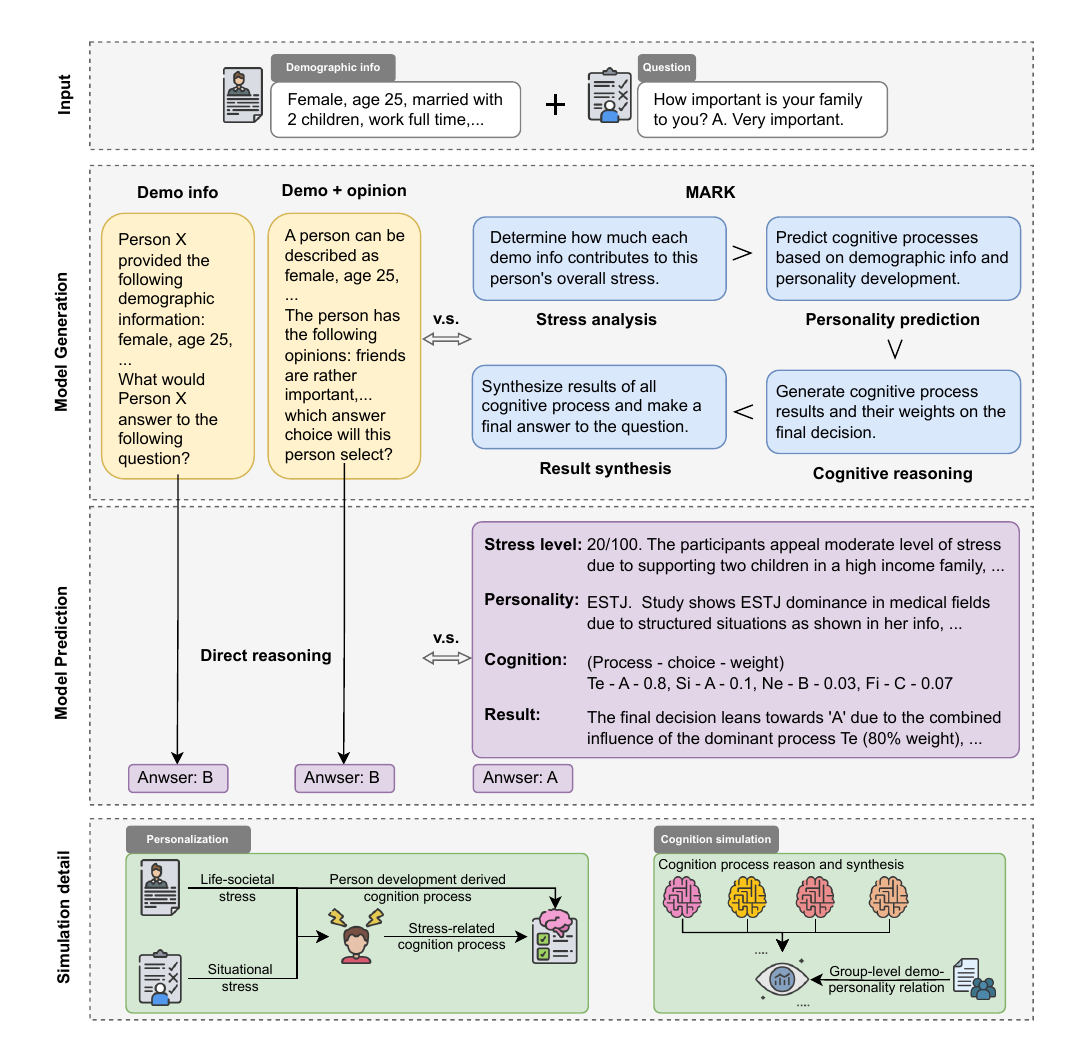}

    \caption{The MARK framework enhances cultural value survey simulation via multi-stage reasoning over MBTI personality theory with life-situational stress analysis, demographic-based group-level personality prediction, and weighted cognitive imitation. These enhancements \sethlcolor{lightgreen}\hl{increase simulation accuracy, decision interpretability, and response personalization}.
    }
    \label{fig:prob_fig}
\end{figure}

As illustrated on the left side of Figure \ref{fig:prob_fig}, solutions like demographic augmentation \citep{kwok2024evaluating, zhao-etal-2024-worldvaluesbench} and opinion insertion \citep{hwang-etal-2023-aligning} were developed to improve model simulation responses in cultural value surveys. While these studies have primarily relied on prompting that uses personal demographic information (shown as "demo info") as input to the model, models still struggle with zero-shot personalization, and prediction interpretation is still challenging.

In this paper, we focus on \textbf{improving cultural value survey simulation by incorporating psychological theory} into the modeling of human responses. Our key insight is that human responses to survey questions are significantly influenced by their mental status (like stress) and associated cognitive processes. To capture these influences, we build a \textbf{M}ulti-st\textbf{A}ge \textbf{R}easoning framewor\textbf{K} (\textbf{MARK}) that simulates how different personality types process information and make decisions. 
MARK draws on the MBTI personality type dynamic theory \citep{Mitchell01type} to model cognitive functions and their interactions with stress and personal experience, aiming to achieve accurate and psychologically grounded survey simulations than previous methods that primarily rely on demographic information. 

The process begins with a \textbf{stress analysis} to evaluate societal pressures based on demographic data, which generates sociodemographic prompts that enhance the simulation. Following this, a two-step \textbf{personality prediction} is carried out to infer cognitive functions from these descriptions, utilizing MBTI personality theory. During the survey simulation, a \textbf{cognitive reasoning} stage mimics human thought processes under both life- and situational-stress, influenced by the interplay between human traits and specific questions. Finally, the process concludes with a \textbf{synthesis} stage that reviews the reasoning outcomes to produce the final answer.

To reduce personality assumption bias, we evaluate our framework on a clustered version of the World Values Survey (see Appendix \ref{sec:cls_ana}). Experiments show that MARK can outperform existing baselines by 10\% and reduce the divergence between model predictions and human preferences. Further analysis reveals the robustness of our multi-stage reasoning framework - \textit{MARK performs stably during simulation even with uncertain personalities}. The fidelity of our framework suggests that incorporating personality-driven cognitive processes is a promising direction for improving survey simulation accuracy while providing deeper insights into human decision-making patterns.

Our contributions are threefold:
\begin{enumerate}
    \item We designed a multi-stage reasoning framework with personality cognition called MARK to simulate the human decision process on value surveys.
    \item Experiments on simulations show that MARK outperforms other baselines by presenting higher accuracy and generalizes consistently across the U.S. and Chinese cultures.
    \item Further analysis suggests that the system performs robustly with uncertain personalities, benefiting from the multi-stage design.
\end{enumerate}

\section{Related Work}
In the field of AI for social science research, researchers have increasingly explored the use of LLMs to \textbf{simulate human responses to survey questions} \citep{argyle2023out, cao-etal-2025-specializing, LIU25Towards}. 
Researchers investigate LLMs' predictive ability of simulating cross-cultural variations \citep{kwok2024evaluating, cao-etal-2023-assessing, alkhamissi-etal-2024-investigating}, value diversity \citep{santurkar2023opinionslanguagemodelsreflect, zhao2024world}, and modeling of complex value systems \citep{benkler2023assessing, arora-etal-2023-probing}. 

Moreover, LLMs also show promise in replicating human samples across various fields, including economics \citep{horton2023large}, psychology \citep{aher2023using}, and other domain tasks \citep{Luo2025ASO}. However, these studies also highlight the limitations of LLMs in capturing the intricacies of human behavior, particularly in nuanced decision-making situations \citep{bisbee2023synthetic}.

To improve simulation accuracy, researchers have investigated various prompting strategies \citep{mou2024individualsocietysurveysocial}. A key approach has been the integration of demographic information into LLMs to enhance zero-shot simulations \citep{deshpande-etal-2023-toxicity}, known as \textbf{sociodemographic prompting}. However, studies have found that LLMs tend to produce stereotypes and biases under straightforward demographic prompting conditions \citep{cheng-etal-2023-marked}, complicating efforts to simulate human behaviors with greater precision.

To address these challenges, researchers have explored more sophisticated prompting methods. \citet{hwang-etal-2023-aligning} proposed incorporating user opinions to better reflect specific group perspectives. \citet{zhou-etal-2023-cultural} demonstrated that including cultural background information could enhance model transferability. \citet{cao-etal-2025-specializing} shows that fine-tuning on first token probabilities can better align simulated and real response distributions. However, recent studies by \citet{li2024culturegenrevealingglobalcultural} and \citet{shen-etal-2024-understanding} reveal that current methods still fall short in simulating preference diversity and the intricate decision-making processes characteristic of human behavior. This limitation underscores the need for more personalized and psychologically grounded approaches to survey simulation.

\section{MARK: Multi-stAge Reasoning frameworK}
\begin{figure*}
    \centering
    \includegraphics[width=\linewidth]{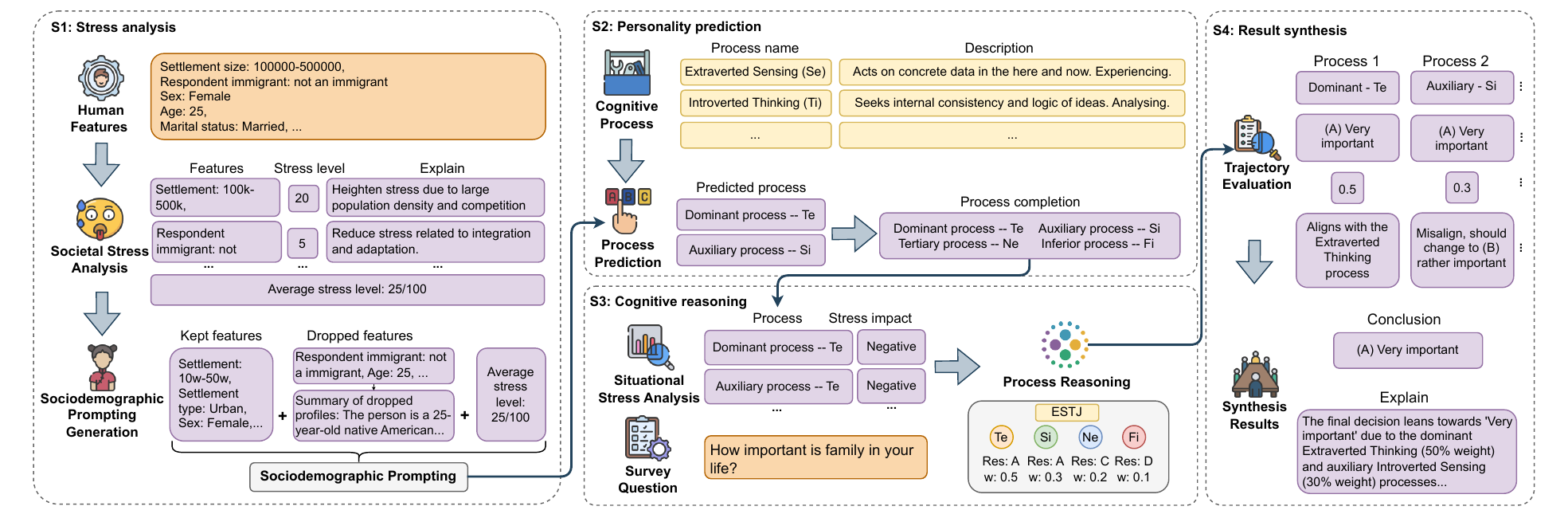}
    \caption{Overview of the proposed multi-stage reasoning framework for simulating individual cognition in survey responses. The framework comprises four stages: 1) \textbf{Stress Analysis}—evaluates life-societal pressures to generate stress levels and a summarized human profile; 2) \textbf{Personality Prediction}—predicts cognitive functions and personality traits based on the profile; 3) \textbf{Cognitive Reasoning}—assesses stress impact of specific questions, conducts reasoning using predicted functions, and assigns weights to finalize answers; and 4) \textbf{Synthesis}—reviews reasoning history, validates results, and synthesizes them into a cohesive response. Color-coded data sources: \sethlcolor{lightorange}\hl{Orange} boxes indicate survey data, \sethlcolor{lightyellow}\hl{yellow} boxes represent function call outputs, and \sethlcolor{lightpurple}\hl{purple} boxes denote text generated by LLMs.}
    \label{fig:mark_pipeline}
\end{figure*}

The Myers-Briggs Type Indicator \citep[MBTI]{Clawson08Myers} is a widely used personality assessment tool that categorizes individuals into 16 distinct personality types. Each personality type is characterized by the interactions between four cognitive functions that describe the mental processes in the order of how we naturally access them and their development—Dominant, Auxiliary, Tertiary, and Inferior functions \citep{Mitchell01type}.

The Dominant Function is the most developed, conscious, and comfortable aspect of an individual's personality, serving as their core and primary cognitive function. The Auxiliary Function supports and balances the Dominant Process, aiding in perception, judgment, and the balance between Extraversion and Introversion. The Tertiary Function is less competent and comfortable than the Dominant and Auxiliary, typically activated under stress for decision-making. The Inferior Function is the least developed, emerging under extreme stress and potentially causing individuals to feel out of control.
(Details in Appendix \ref{app:cfun})

To capture how these personality dynamics influence survey responses, we develop MARK, a multi-stage reasoning framework that simulates human decision-making through the lens of personality type dynamics. Our key insight is that personality traits and their associated cognitive functions significantly impact how individuals process information and respond to survey questions, particularly under varying levels of stress.

As presented in Figure \ref{fig:mark_pipeline}, MARK organizes four major reasoning stages: stress analysis, personality prediction, cognitive reasoning, and synthesis. We provide a brief summary of stage functionality in Table \ref{tab:summed_p} (detailed in Appendix \ref{app:prompts}).

\paragraph{\textit{Stage 1}: Stress analysis} The MBTI-type dynamic theory emphasizes the impact of stress on triggering different thinking functions and inducing the positive or negative reasoning pattern of the process \citep{Clark02stress}. Before the personality reasoning of each human subject, we ask the model to evaluate the features recorded in surveys and predict the stress level of each feature $S_{c_i}$ (0 to 100). We use the average stress level of all the features as the overall stress level $S_{\text{overall}}$ of the participant. 

As personalities develop throughout life, the demographic information captured in our surveys reflects this development and provides insights into possible personality types. However, identifying which features have the greatest impact on simulation results presents a challenge. To address this, we developed an automatic sociodemographic prompting method that preserves features associated with high-stress levels and key personality interactions. Our method analyzes both relationship dependencies between features and feature complexity to maintain the integrity of preserved features while summarizing less critical ones. For any demographic information not retained, the model generates profile descriptions that maintain the overall representation of the human subject.

\begin{table*}[]
\resizebox{\textwidth}{!}{%
\begin{tabular}{l|llll}
\toprule
\textbf{Stage} &
  \textbf{Step} &
  \textbf{Input} &
  \textbf{Key Function} &
  \textbf{Output} \\ \midrule
\multirow{2}{*}{\textbf{1. Stress Analysis}} &
  Feature Evaluation &
  Human features &
  Assign stress levels (0-100) to each feature &
  Features with stress scores \\
 &
  Profile Filtering &
  Stress-scored features &
  Retain high-stress features, summarize dropped ones &
  Filtered feature profile \\ \midrule
\textbf{2. Personality Prediction} &
  Process Selection &
  Filtered features + stress level &
  Select MBTI cognitive processes &
  D-A-T-I processes \\ \midrule
\multirow{2}{*}{\textbf{3. Cognitive Reasoning}} &
  Stress Impact Assessment &
  MBTI processes + survey questions &
  Determine if stress triggers negative aspects &
  Processes with stress impact \\
 &
  Multi-Process Reasoning &
  Stress-impacted processes + survey &
  Generate reasoning for each process with weights &
  Weighted reasoning results \\ \midrule
\textbf{4. Synthesis} &
  Result Integration &
  All process reasoning + weights &
  Evaluate consistency and synthesize final answer &
  Final survey response \\ \bottomrule
\end{tabular}%
}
\caption{Overview of the four-stage multi-agent reasoning framework for human value preference simulation.}
\label{tab:summed_p}
\end{table*}

\begin{figure}
    \centering
    \includegraphics[width=.9\linewidth]{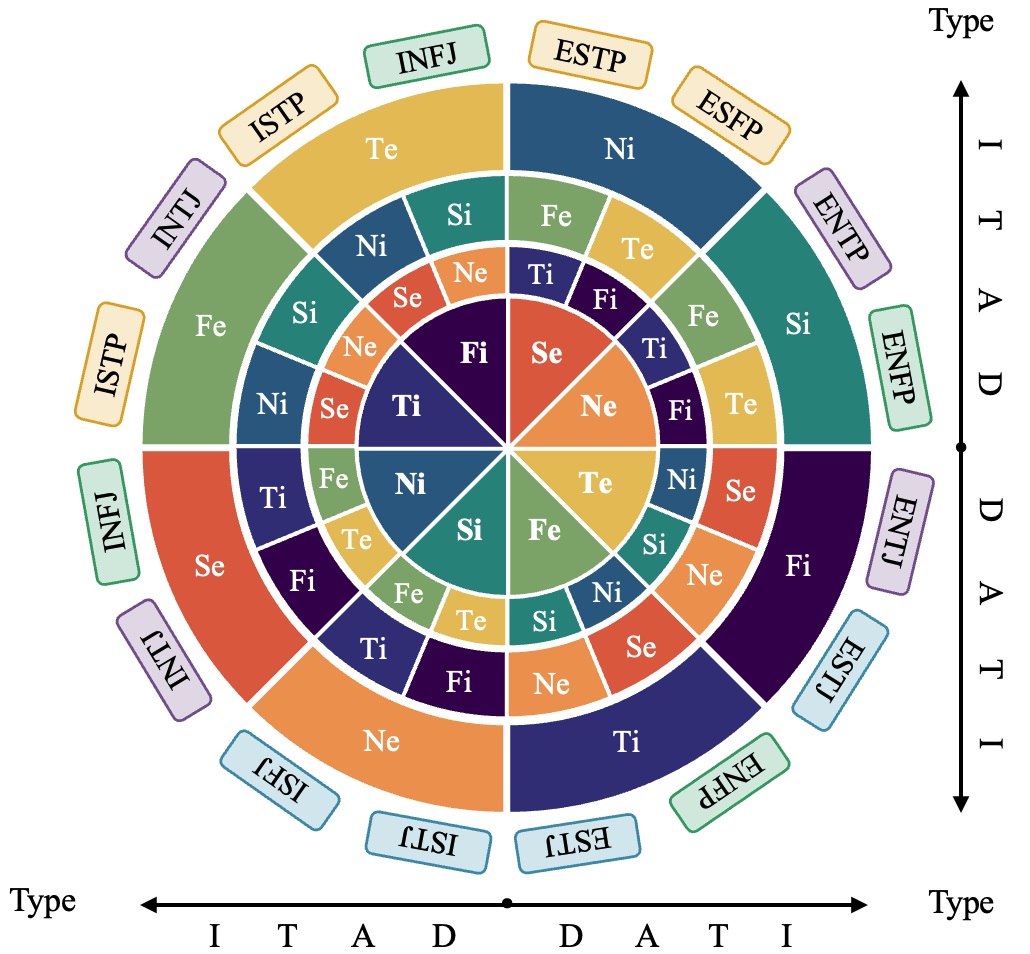}
    \caption{Personality cognition compass: how cognitive functions merge to personality types. The "DATI" represents "Dominant", "Auxiliary", "Tertiary", and "Inferior" functions. Once the first two functions are selected, the corresponding personality type is set.}
    \label{fig:funtotype}
\end{figure}

\paragraph{\textit{Stage 2}: Personality prediction}
The MBTI type dynamics theory \citep{Mitchell01type} establishes a comprehensive pattern of how cognitive functions form different personalities. The personality cognition compass (Figure \ref{fig:funtotype}) shows that personality types can be determined once the first two functions are selected.

While the MBTI assessment traditionally relies on self-reported behaviors, demographic factors serve as \textbf{indirect proxies} for environmental pressures that systematically influence personality development through mediating pathways \citep{myers1998mbti, Rentfrow2013Divided}. Rather than assuming direct demographic-personality mappings, our framework \textbf{leverages demographic features to parameterize contextual priors} reflecting how occupational, educational, and geocultural environments may shape cognitive function preferences over time. This approach provides greater mechanistic sophistication than direct demographic-to-attitude prediction by modeling personality as an intermediate latent construct influenced by environmental factors.

During personality prediction, we prompt the model to predict dominant cognitive functions using demographic information as environmental context. The system selects the primary function, then applies type dynamics theory to identify auxiliary function candidates, reducing prediction errors. Once these two functions are determined, the remaining functions are derived to form the complete personality type. The prediction algorithm is detailed in Appendix \ref{sec:person_predict}.

\paragraph{\textit{Stage 3}: Cognitive reasoning} 
The MBTI cognitive process exhibits two reasoning patterns influenced by societal and situational stress \citep{Clark02stress}: 1) Under low situational stress, life-societal stress predominates, leading to normal cognitive functioning; 2) Under high situational stress, often triggered by past trauma, individuals experience heightened stress and overuse of cognitive processes, similar to Post-traumatic Stress Disorder \citep[PTSD]{fang20impact}. In this stage, we evaluate the interaction of these stress types to determine if individuals are vulnerable to cognitive overload. Detailed descriptions of normal and overused functions are provided in Appendix \ref{app:cfun}.

For each thinking process, we assign stress impact levels and ask the model to reason about cognitive status and generate value preferences. To determine the confidence in each function's result, we also assign influence weights for each process, which are used in the synthesis stage.

\paragraph{\textit{Stage 4}: Synthesis} 
During synthesis, we collect and document the complete reasoning trajectory. The model reviews these to ensure consistency with the explanations and process descriptions. If it finds misalignments, it rewrites the results to minimize hallucinations and verifies that appropriate weights are assigned to each process in the MBTI hierarchy.

Following this evaluation, the model synthesizes the reasoning results based on weights and selects the most appropriate survey response option. To enhance explainability, it provides a brief explanation of how weights influenced the final decision.

% \paragraph{\textit{Stage 5}: Cultural Modification} Recently, \citet{Williams24culture} highlighted the interaction between personality and culture. Based on their findings, we introduce the role of cultural context in the final step of MARK reasoning. We ask the model to review simulated preferences and assess their alignment with cultural values. If aligned, the agent keeps the original conclusion; if misaligned, it should adjust its conclusions to better reflect cultural values. It’s important to note that individual experiences may differ from general cultural orientations, so we instruct the model that the significance of cultural values should be reduced unless individuals fit cultural stereotypes.
\section{Experiment}
\begin{table*}[]
\centering
\resizebox{\textwidth}{!}{%
\begin{tabular}{l|rrrr|rrrr|rrrr}
\toprule
\multicolumn{1}{c|}{\textbf{Model}} &
  \multicolumn{4}{c|}{\textbf{GLM-4-air}} &
  \multicolumn{4}{c|}{\textbf{GPT-4o}} &
  \multicolumn{4}{c}{\textbf{Doubao-1.5-pro}} \\ 
  \midrule \midrule
\multicolumn{13}{c}{\textit{Global Distribution}} \\
  \midrule \midrule
\textbf{Avg. Metrics} &
   &
  \textbf{1-JSD} $\uparrow$ &
  \textbf{EMD} $\downarrow$ &
  \textbf{$\kappa$} $\uparrow$ &
   &
  \textbf{1-JSD} $\uparrow$ &
  \textbf{EMD} $\downarrow$ &
  \textbf{$\kappa$} $\uparrow$ &
   &
  \textbf{1-JSD} $\uparrow$ &
  \textbf{EMD} $\downarrow$ &
  \textbf{$\kappa$} $\uparrow$\\
    \midrule
    \textbf{$\text{Random selection}^{\dag}$} &
   &
  0.3319 &
  0.2848 &
  0.0000 &
   &
  0.3319 &
  0.2848 &
  0.0000 &
   &
  0.3319 &
  0.2848 &
  0.0000 \\
\textbf{Without Demo.} &
   &
  0.4492 &
  0.1398 &
  0.0125 &
   &
  0.4421 &
  0.1177 &
  0.0138 &
   &
  0.3788 &
  0.1455 &
  0.0112 \\
\textbf{\citet{durmus2024measuringrepresentationsubjectiveglobal}} &
   &
  0.2664 &
  0.1466 &
  0.0142 &
   &
  0.4411 &
  0.0981 &
  -0.0071 &
   &
  0.3544 &
  0.1280 &
  0.0116 \\
\textbf{\citet{LIU25Towards}} &
   &
  0.4371 &
  0.0490 &
  0.0118 &
   &
  0.4620 &
  0.0211 &
  0.0113 &
   &
  0.5258 &
  0.0218 &
  0.0104 \\
\textbf{Demo.+Ideo.} &
   &
  0.4336 &
  0.1041 &
  0.0011 &
   &
  0.6404 &
  0.0987 &
  0.0112 &
   &
  0.5520 &
  0.1011 &
  0.0180 \\
\textbf{Demo.+Ideo.+Opinion} &
   &
  0.5425 &
  0.1112 &
  0.0180 &
   &
  0.6449 &
  0.0981 &
  0.0097 &
   &
  0.5695 &
  0.0995 &
  0.0198 \\
\textbf{\citet{zhao-etal-2024-worldvaluesbench}} &
   &
  0.4267 &
  0.1280 &
  0.0024 &
   &
  0.6174 &
  0.1113 &
  0.0111 &
   &
  0.3974 &
  0.1265 &
  0.0123 \\
    \midrule \textbf{MARK (Ours)} &
   &
  \textbf{0.5963} &
  \textbf{0.1026} &
  0.0127 &
   &
  0.6438 &
  0.0985 &
  \textbf{0.0128} &
   &
  \textbf{0.6111} &
  0.1141 &
  0.0123 \\
    \midrule \midrule
\multicolumn{13}{c}{\textit{Sampled Distribution}} \\
  \midrule \midrule
\textbf{Avg. Metrics} &
  \textbf{ACC} (\%) &
  \textbf{1-JSD} $\uparrow$ &
  \textbf{EMD} $\downarrow$ &
  \textbf{$\kappa$} $\uparrow$ &
  \textbf{ACC} (\%) &
  \textbf{1-JSD} $\uparrow$ &
  \textbf{EMD} $\downarrow$ &
  \textbf{$\kappa$} $\uparrow$ &
  \textbf{ACC} (\%) &
  \textbf{1-JSD} $\uparrow$ &
  \textbf{EMD} $\downarrow$ &
  \textbf{$\kappa$} $\uparrow$ \\
    \midrule
\textbf{Demo.+Ideo.} &
  25.49 &
  0.3539 &
  0.0741 &
  0.02 &
  32.30 &
  0.4075 &
  0.0755 &
  0.09 &
  30.75 &
  0.4563 &
  0.0685 &
  0.02 \\
\textbf{Demo.+Ideo.+Opinion} &
  25.06 &
  0.4313 (0.6) &
  0.0669 &
  0.11 &
  33.07 &
  0.4069 &
  0.0739 &
  0.09 &
  31.45 &
  0.4723 &
  0.0703 &
  0.08 \\
\textbf{\citet{zhao-etal-2024-worldvaluesbench}} &
  26.98 &
  0.3814 &
  0.0452 &
  0.05 &
  36.96 &
  0.4654 &
  0.0610 &
  0.12 &
  24.23 &
  0.3584 &
  0.0364 &
  0.05 \\
    \midrule 
\textbf{MARK (Ours)} &
  \textbf{33.69} &
  \textbf{0.4348} &
  0.0887 &
  \textbf{0.15} &
  \textbf{38.11} &
  \textbf{0.4879} &
  0.0826 &
  \textbf{0.15} &
  \textbf{46.98} &
  \textbf{0.5195} &
  0.0561 &
  \textbf{0.09}\\
  \bottomrule
\end{tabular}%
}

\caption{MARK simulation performance on U.S. social‐survey data, evaluated under both global‐distribution and sampled‐distribution settings. `\textit{Avg.}' denotes the overall mean metric values, (value) denotes the p-value of results between baseline and MARK larger than 0.05. MARK shows improvements on sampled distributions by achieving the \textbf{highest simulation accuracy} while \textbf{maintaining low divergences}, with most improvements being statistically significant. It also shows \textbf{generalization to global distributions} with unseen demographics.}
\label{tbl:chara_res}
\end{table*}
\subsection{Dataset}
\label{sec:data}
Our simulation samples are drawn from the $7^{th}$ wave of the World Values Survey, a comprehensive global study that examines changing values and their influence on social and political life. For evaluation purposes, we utilize survey data from the United States and China as case studies.

\paragraph{Setting 1: Global Distribution}
To evaluate the generalization of MARK on responses over the whole survey population in specific cultural backgrounds, we also compare the model-predicted responses based on sampled demographic information, presented in setting 2, with the distribution divergence and agreement with the whole survey population (\textit{Global Distribution} setting).

\paragraph{Setting 2: Sampled Distribution}
To encompass a broader range of human features, we examine 30 features detailed in the survey, such as continent, residential area, and education level \citep{zhao-etal-2024-worldvaluesbench}. To reduce assumption bias for individual data, representatives are selected through a group-level clustering approach (\textit{Sampled Distribution} setting) reflecting the diversity of demographics.

We conducted a group-level demographic-based clustering analysis in the following steps: 1) Numerically transform and normalize the demographic feature attributes for human subjects from the social survey. 2) Implement the Silhouette Analysis \citep{oncevay-etal-2020-bridging} to determine cluster optimality and apply the K-means clustering.

Based on analysis (see Appendix \ref{sec:cls_ana}), we conducted the configuration of 20 clusters and sampled 20 subjects from each cluster to represent U.S. culture, utilizing a dataset of 102,800 QA pairs. To evaluate the generalizability of our method, we applied the same clustering approach to the Chinese version of the survey and selected the cluster centroids as our simulation targets. Ultimately, we obtained 20 subjects for the simulation, using a dataset of 4,340 QA pairs.

Moreover, we augmented the World Values Survey with the assumed personalities (oracle personality) of human participants to study the robustness of our framework. (See Appendix \ref{sec:data_aug})
\begin{figure*}
    \centering
    \subfloat[Performance of GLM-4-air model.]{
    \includegraphics[width=0.45\linewidth]{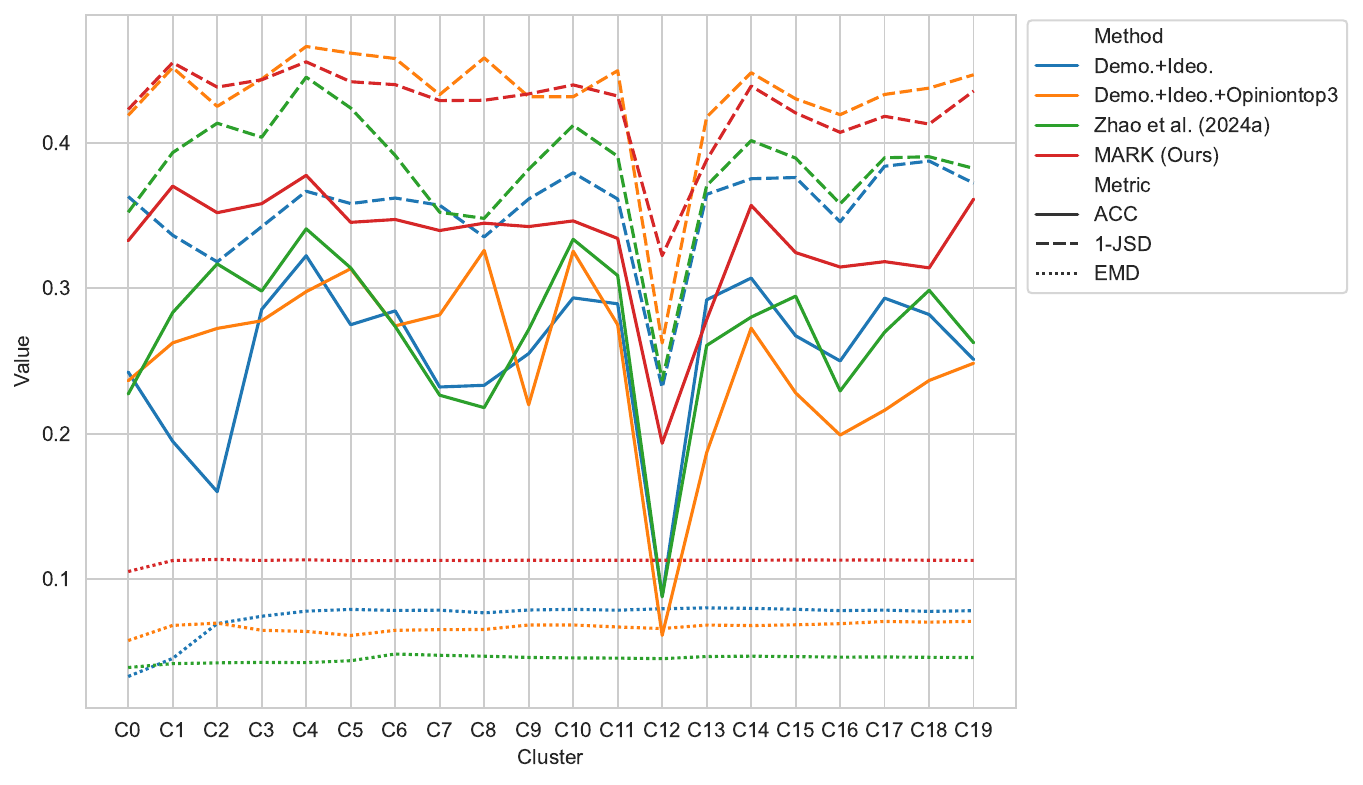}
    }
    \subfloat[Performance of Doubao-1.5-pro model.]{
    \includegraphics[width=0.45\linewidth]{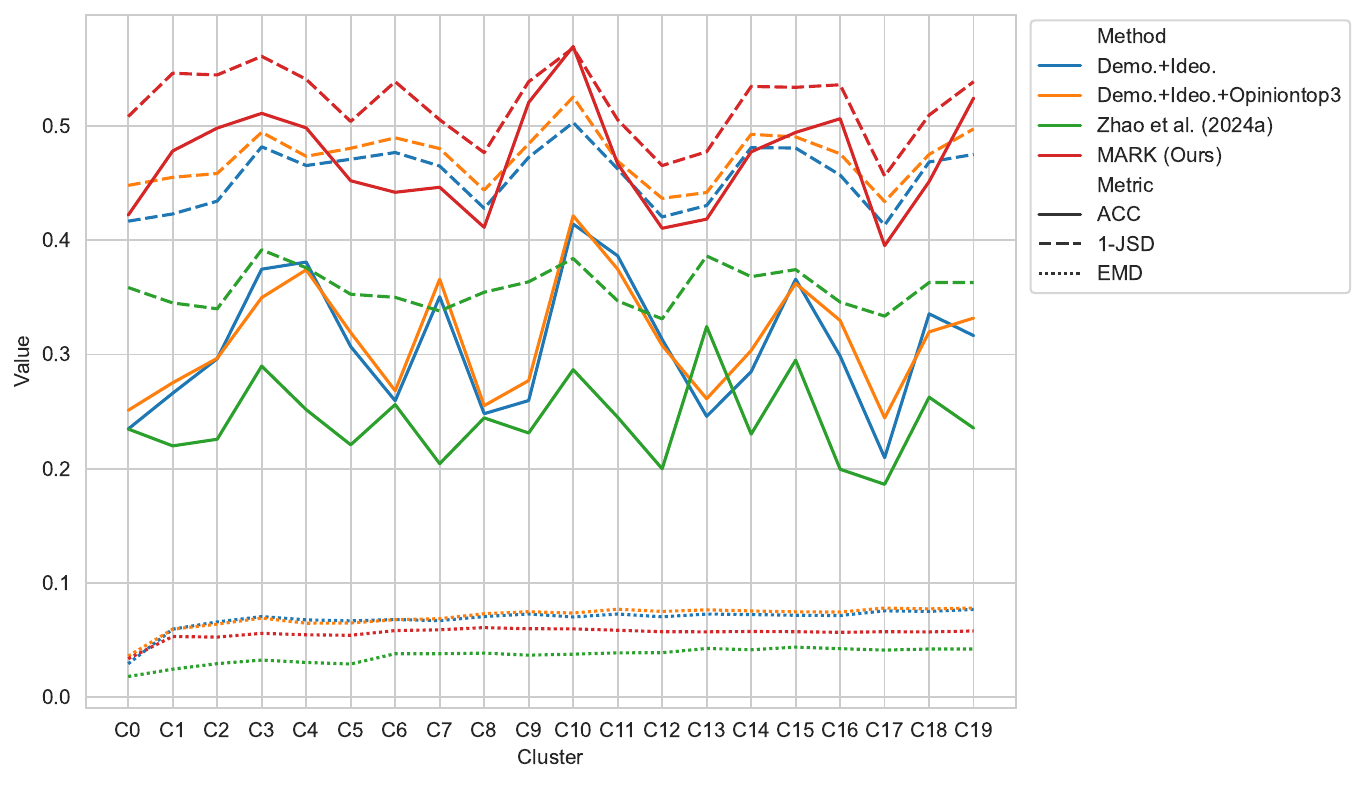}
    }
    \caption{Group-level performance of survey simulation on U.S. culture in the \textit{Sampled Distribution} setting. MARK shows significant improvement over other baseline methods on most clusters in terms of ACC and 1-JSD.}
    \label{fig:full_res}
\end{figure*}
\subsection{Baselines} 
We introduce several baseline methods without personal demographic info by \citet{durmus2024measuringrepresentationsubjectiveglobal} to investigate the generalization of MARK on the \textit{Global Distribution} setting:

\textbf{$\text{Random selection}^{\dag}$} We present a random selection as the lower bound performance for the task. Given that random selection is independent of any model, the performance outcomes remain consistent across all models. To facilitate improved comparison, these results are reiterated in the accompanying table.

\textbf{Without Demo.} \citep{durmus2024measuringrepresentationsubjectiveglobal} uses multiple-choice survey questions originally written without any indication of the personal demographic information (DP setting in their paper).
\textbf{Nation Info Only} \citep{durmus2024measuringrepresentationsubjectiveglobal, cao-etal-2025-specializing, LIU25Towards} asks LLMs to predict how someone from the U.S. or China would respond to these questions. We chose two standard baseline prompt as this kind of basline: \textbf{\citet{durmus2024measuringrepresentationsubjectiveglobal} (CP setting)} and \textbf{\citet{LIU25Towards}}.

We also select three baseline methods presented by \citet{hwang-etal-2023-aligning} and \citet{zhao-etal-2024-worldvaluesbench} using the demo. info to investigate the alignment of LLMs with individual user preferences, focusing on modeling user demographics, ideologies, and past opinions to predict future responses:

\textbf{Demo.+Ideo.} \citep{hwang-etal-2023-aligning} combines demographic information and ideological inclinations of a user to predict their responses to survey questions.
\textbf{Demo.+Ideo.+Opiniontop3} \citep{hwang-etal-2023-aligning} combines demographic information, ideological inclinations, and the top 3 most relevant opinions to predict user responses.
\textbf{\citet{zhao-etal-2024-worldvaluesbench}} focuses on three demographic variables (i.e., the continent, resident area, and education level) to indicate a human and asks the model to predict their preferences on survey questions.

\subsection{Models} All the experiment is conducted with GLM-4-air \citep{glm2024chatglmfamilylargelanguage}, GPT-4o (gpt-4o-2024-08-06) \citep{gpt-4o}\footnote{Due to budget limitations, we only conduct simulations on cluster centers on GPT-4o model.}, and Doubao-1.5-pro\footnote{\url{https://team.doubao.com/en/direction/llm}} via API\footnote{We also test on Llama-3.1-7B/13B, and found they are limited in conducting cognitive simulations, likely due to the complexity of instructions and human features.}, and the embedding model used by Demo.+Ideo.+Opiniontop3 is embedding-v3 by ZHIPU AI\footnote{\url{http://open.bigmodel.cn/dev/howuse/model}}. The experiments for the U.S. and Chinese simulations were completed with a budget of \$1000. The generation temperature is 0.9, and the max token length is 4096.

\subsection{Metrics} We evaluate the simulation performance via the preference distribution gathered from the responses and prediction accuracy with exact text matching.

\paragraph{Response accuracy (\textit{ACC})} ACC measures how well the model predicts the exact responses compared to the actual labels. We obtain ACC by calculating the percentage of the correct responses of all responses in the cluster and all simulation targets.
Higher ACC indicates better alignment.

\paragraph{Distribution divergence (\textit{1-JSD, EMD})} 1-JSD and EMD metrics evaluate how well the distribution of the model's predictions aligns with the actual human distribution. (1) 1-JSD, known as 1 minus the Jensen-Shannon divergence, implemented by \citet{durmus2024measuringrepresentationsubjectiveglobal}, is a symmetric measure of the similarity between two probability distributions; (2) the Earth Mover Distance \cite[EMD]{rubner1998metric}, known as the Wasserstein distance, quantifies the minimum amount of work required to transform one distribution into another.

Particularly for the Sampled Distribution setting, we collect responses from each cluster per question, then plot distributions for human participants and model predictions within the cluster. Then, we calculate the 1-JSD and EMD score over these distributions. We took the mean values across all clusters as average scores. 
Lower EMD and higher 1-JSD values indicate better alignment.

\paragraph{Cohen’s $\kappa$} utilized by \citep{badshah24reference} measuring the percentage of agreements on the group-level voting. We calculate the agreement of model predictions and human responses using Cohen’s $\kappa$. Higher agreement shows better performance.

\begin{figure*}
    \centering
    \subfloat[Performance of GLM-4-air model.]{
    \includegraphics[width=0.45\linewidth]{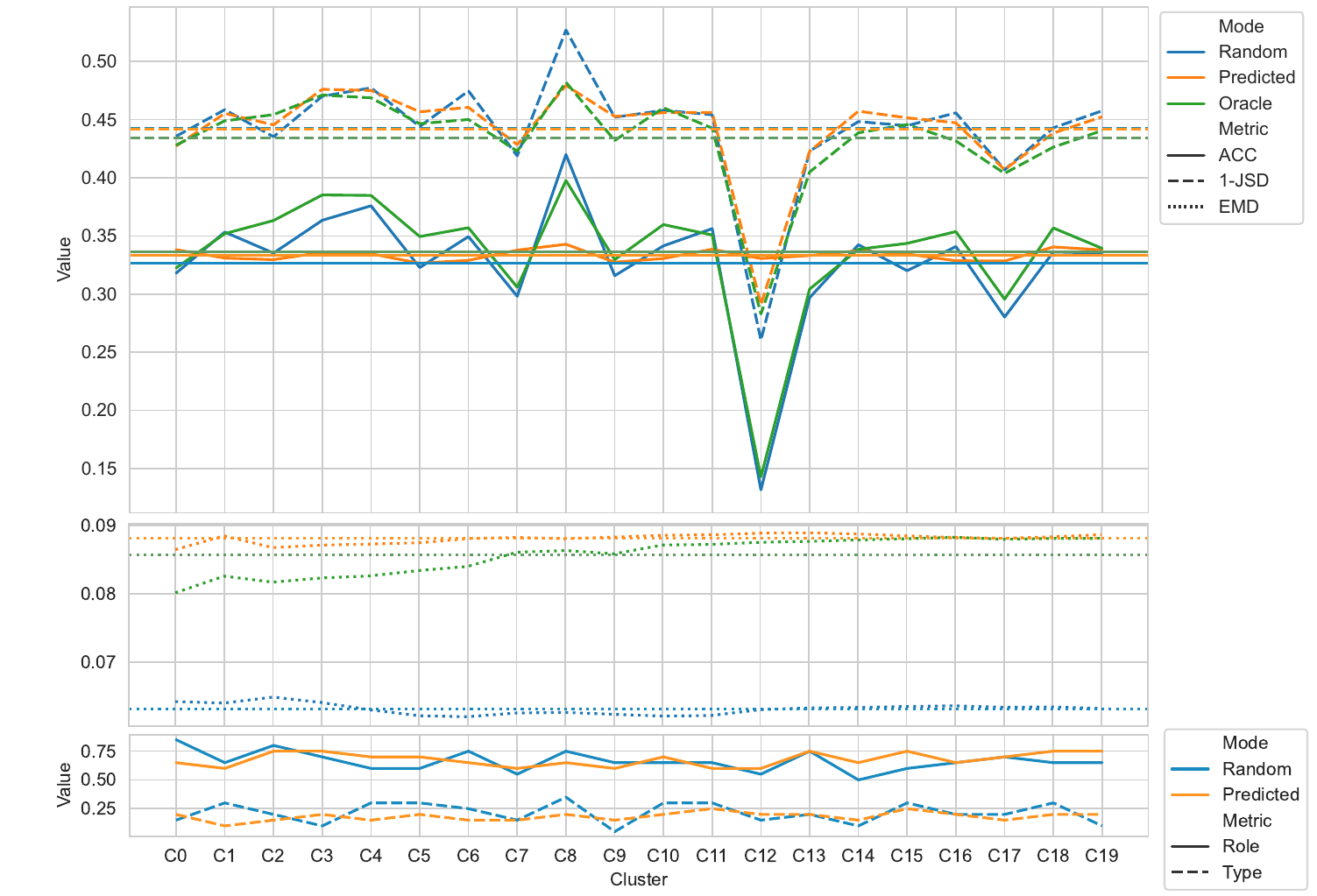}
    }
        \subfloat[Performance of Doubao-1.5-pro model.]{
    \includegraphics[width=.45\linewidth]{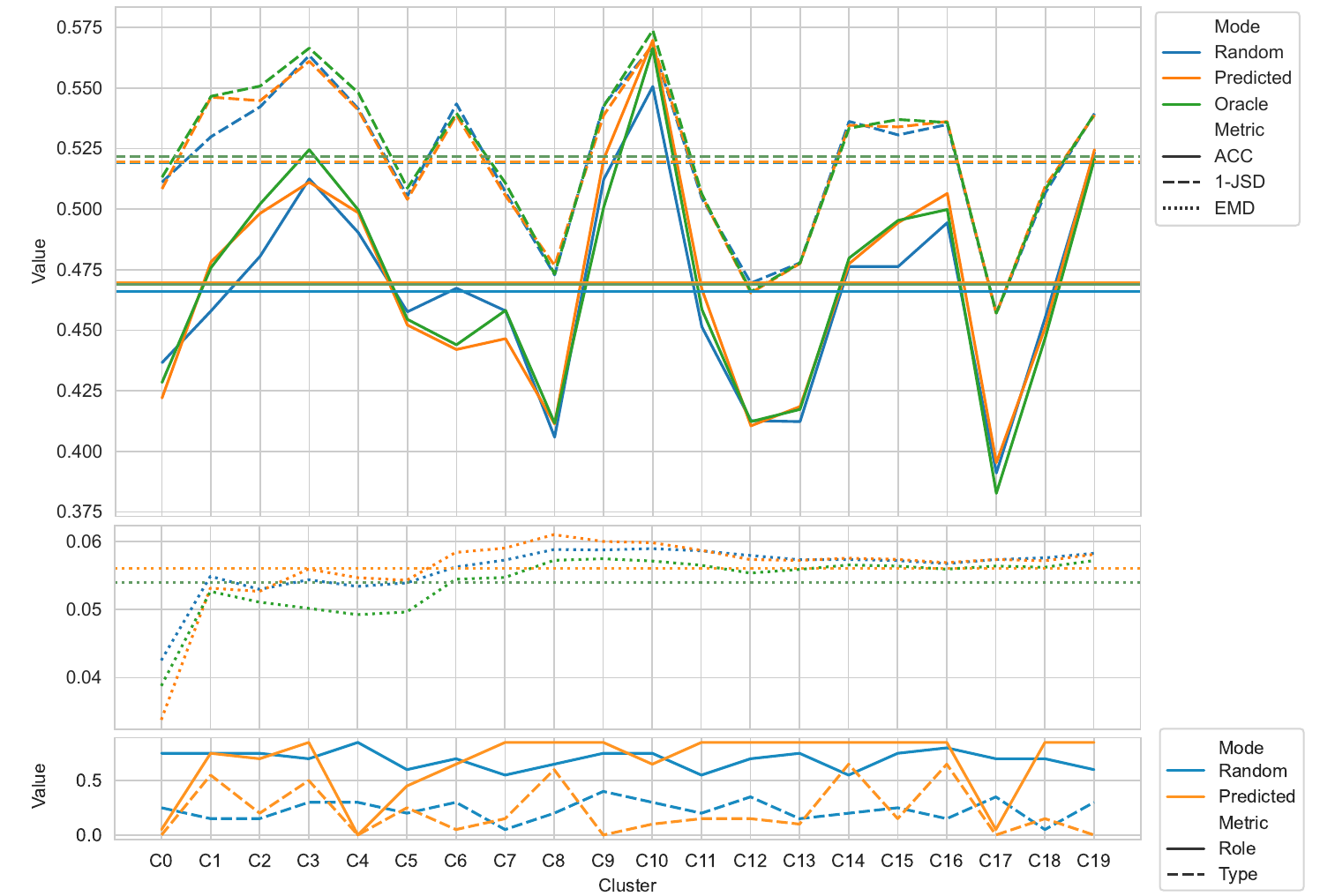}
    }
    \caption{Group-level results of survey simulation on U.S. culture with various personality strategies in the \textit{Sampled Distribution} setting. The horizontal line indicates the average score of each metric. MARK shows robust performance across strategies and exceeds with oracle personalities.}
    \label{fig:robustness}
\end{figure*}

\section{Results}
This section presents the experimental results and examines the impact of incorporating personality traits for simulation. Additionally, we evaluate the system's robustness, generalization on the Chinese culture, and significance of each component in the MARK system through ablation studies conducted on best- and worst-performing models.

\subsection{Generalization to Global Distribution}
Table \ref{tbl:chara_res} (top) reports results when each model’s aggregate predictions—conditioned solely on demographics (or nothing)—are compared to the full U.S. survey population histogram. MARK shows competitive performance: it achieves the competitive 1–JSD (0.5963 on GLM‑4‑air; 0.6438 on GPT‑4o; 0.6111 on Doubao‑1.5‑pro) and the lower EMD (0.1026, 0.0985, 0.1141, respectively), indicating its simulated population closely mirrors the true preference distribution. Cohen’s $\kappa$ remains at or above the level without demo. info and nation info only prompts, underscoring robust group‐level agreement even under sparse conditioning.

\subsection{Sampled‑Distribution Performance}
Table \ref{tbl:chara_res} (bottom) summarizes each model’s accuracy (ACC), distributional alignment (1–JSD, EMD), and category‑agreement (Fleiss’$\kappa$) when predicting the held‑out human responses from our sampled participant pool. Across all three base LLMs (GLM‑4‑air, GPT‑4o, Doubao‑1.5‑pro), MARK achieves the highest ACC, improving over the strongest “Demo+Ideo+Opinion” baseline by 4.3\% on GLM‑4‑air, 8.9\% on GPT‑4o, and a striking 15.5\% on Doubao‑1.5‑pro. These gains are statistically significant (p<0.05), \textbf{demonstrating its robustness in predicting exact responses}.

In tandem with higher accuracy, MARK also yields the best distributional fidelity: it attains the largest 1–JSD scores (indicating minimal Jensen–Shannon divergence) and competitive EMD values. For example, on GLM‑4‑air, MARK’s 1–JSD of 0.4348 and EMD of 0.0887 outperform both the pure demographic baselines (1–JSD$\simeq$0.35–0.43, EMD$\simeq$0.045–0.074) and \citet{zhao2024world}. Finally, MARK’s Fleiss’ $\kappa$ (0.15) more than triples the “Demo+Ideo” baseline (0.02), demonstrating that its categorical vote proportions align substantially better with the actual human label distributions.

Further analysis in Figure \ref{fig:full_res} shows that MARK presents different sensitivity over certain groups of humans with similar demographic background: 
For example, GLM-4-air shows a \textit{strong performance in demographics of married working full-time jobs} like in $C_0$ (34.28\%) and $C_5$ (34.59\%). 
And Doubao-1.5-pro \textit{excels in demographics of skilled workers for private business} as in clusters $C_{10}$ (56.96\%) and $C_{19}$ (52.45\%), where it outpaces competitors by large margins.

These superior performances of MARK can be attributed to its integration of MBTI personality cognition, which enhances its ability to simulate human-like responses in social surveys. By incorporating cognitive traits, MARK captures nuanced behavioral patterns of the demographic information and value preferences. This cognitive augmentation allows MARK to better align with real-world survey responses, leading to higher accuracy and demonstrating the importance of psychological dimensions in social survey simulations.

\subsection{Framework Robustness}

\begin{table*}[]
\centering
\subfloat[Result on Chinese prompts.]{
\label{tbl:en_prompt}
\resizebox{0.48\linewidth}{!}{%
\begin{tabular}{l|ccc}
\toprule
Method & avg\_ACC & avg\_1-JSD↑ & avg\_EMD↓ \\ \midrule \midrule
\multicolumn{4}{c}{GLM-4-air} \\ \midrule
Demo.+Ideo. & 25.48\% & 0.4206 & 0.0848 \\
Demo.+Ideo.+Opiniontop3 & 35.21\% & 0.5293 & 0.0861 \\
\citet{zhao-etal-2024-worldvaluesbench} & 27.07\% & 0.4169 & 0.0932 \\
\textbf{MARK (Ours)} & 31.58\% & \textbf{0.5704} & \textbf{0.0620} \\ \midrule
\multicolumn{4}{c}{Doubao-1.5pro} \\ \midrule
Demo.+Ideo. & 30.35\% & 0.5007 & 0.0810 \\
Demo.+Ideo.+Opiniontop3 & 30.55\% & 0.5128 & 0.0830 \\
\citet{zhao-etal-2024-worldvaluesbench} & 30.28\% & 0.4407 & 0.0897 \\
\textbf{MARK (Ours)} & \textbf{45.16\%} & \textbf{0.6010} & 0.0861 \\ \bottomrule
\end{tabular}
}
}\vspace{0.5pt}
\subfloat[Results on English prompts.]{
\label{tbl:cn_prompt}
\resizebox{0.48\linewidth}{!}{%
\begin{tabular}{l|ccc}
\toprule
Method & avg\_ACC & avg\_1-JSD↑ & avg\_EMD↓ \\ \midrule \midrule
\multicolumn{4}{c}{GLM-4-air} \\ \midrule
Demo.+Ideo. & 29.47\% & 0.4670 & 0.0877 \\
Demo.+Ideo.+Opiniontop3 & 35.60\% & 0.5339 & 0.0875 \\
\citet{zhao-etal-2024-worldvaluesbench} & 26.24\% & 0.4054 & 0.0997 \\
\textbf{MARK (Ours)} & 31.37\% & \textbf{0.5793} & \textbf{0.0580} \\ \midrule
\multicolumn{4}{c}{Doubao-1.5pro} \\ \midrule
Demo.+Ideo. & 26.98\% & 0.5039 & 0.0888 \\
Demo.+Ideo.+Opiniontop3 & 27.44\% & 0.5244 & 0.0884 \\
\citet{zhao-etal-2024-worldvaluesbench} & 32.24\% & 0.4515 & 0.0923 \\
\textbf{MARK (Ours)} & \textbf{44.08\%} & \textbf{0.6001} & \textbf{0.0839} \\ \bottomrule
\end{tabular}%
}
}
\caption{Experimental results within the Chinese cultural context demonstrate that MARK achieves superior accuracy and the lowest distribution divergence compared to other methods.}
\end{table*}

Due to limited reliable personality data, we use Gemini-2-pro to estimate the likelihood of different personalities selecting survey responses, designating the most probable as the oracle personality\footnote{We treat this kind of personality as an upper-bound performance scenario because this setting has access to the actual survey responses we are attempting to predict.}. Three personality assignment strategies—predicted (\textit{Predicted}), random (\textit{Random}), and oracle (\textit{Oracle})—are integrated into MARK reasoning to assess robustness against personality uncertainty.

As shown in Figure \ref{fig:robustness}, MARK demonstrates strong robustness across all assignment strategies on all models, with simulation accuracy (ACC) and response distribution similarity (1-JSD) remaining \textbf{stable}. For example, Doubao-1.5-pro’s ACC stays at 0.47 (± 0.04–0.05) and 1-JSD between 0.52 and 0.53 across strategies, with similar consistency observed in GLM-4-air.

Personality similarity (role- and type-level) also has \textbf{minimal influence} on simulation performance. Although role-level MBTI similarity is moderately high, type-level alignment is much lower, especially under random assignment. Importantly, clusters with low type similarity do not show reduced ACC or increased divergence. Performance variance across clusters is nearly identical between strategies, and performance curves overlap, indicating stability even with noisy personality data.

MARK’s robustness is attributed to its design: \textit{group-level experimental structure}, \textit{weighted cognitive imitation}, and \textit{stress-informed reasoning}, all of which collectively mitigate the impact of noisy personality inputs and ensure consistent performance across experimental conditions.

\subsection{Generalization to Chinese Culture}

We evaluate the system's generalization to Chinese culture using both Chinese and English prompts on the \textit{Sampled Distribution} setting.

\paragraph{Chinese prompts}
As shown in Table \ref{tbl:cn_prompt}, MARK consistently outperforms baselines in accuracy and response alignment. For Doubao-1.5-pro, MARK achieves 45.16\% ACC, significantly higher than previous benchmarks, and for GLM4, it closely matches the top-performing baseline that leverages additional human opinion data. MARK also records the best distributional metrics (Avg. 1-JSD and Avg. EMD) across both models, underscoring its ability to capture behavioral patterns and align with human preferences.

\paragraph{English prompts}
With English prompts (Table \ref{tbl:en_prompt}), MARK maintains superior performance in the Chinese cultural context. It achieves the highest ACC with Doubao-1.5-pro (44.08\%), and strong distributional alignment for both models (highest Avg. 1-JSD, lowest Avg. EMD).

Notably, MARK’s accuracy and distributional consistency are slightly higher with English prompts, especially for Doubao-1.5-pro, further demonstrating robust generalization to Chinese culture across languages.

\begin{table}[]
\resizebox{\linewidth}{!}{%
\begin{tabular}{c|ccc}
\toprule
\textbf{Method}                   & \textbf{avg\_ACC (\%)} & \textbf{avg\_1-JSD↑} & \textbf{avg\_EMD↓} \\ \midrule
Dominant   process only  & 16.25           & 0.3126               & 0.1856             \\
Auxiliary process only   & 16.21           & 0.3124               & 0.1856             \\
Tertiary process only    & 16.19           & 0.3116               & 0.1856             \\
Inferior process only    & 15.74           & 0.3074               & 0.1851             \\
\textbf{MARK (Full system)} & \textbf{33.69}  & \textbf{0.4348}      & \textbf{0.0887}    \\ \bottomrule
\end{tabular}%
}
\caption{Ablation results within the U.S. cultural context reveal that single-process models are insufficient on their own, underscoring the importance of combining multiple processes to achieve high performance in social survey simulations.
}
\label{tbl:ablation}
\end{table}

\subsection{Ablations} 
We conduct an ablation study to evaluate the impact of dynamic processes within the system on the performance of social survey simulations. 

As shown in Table \ref{tbl:ablation}, the ablation study \textbf{underscores the importance of combining all four processes} in the full MARK system. The dominant process marginally outperforms the auxiliary, tertiary, and inferior processes in predictive metrics, but all single processes are insufficient alone.

This illustrates the critical role of integrating multiple cognitive processes to achieve accurate simulations. Single processes alone are insufficient for producing reliable predictions.

\section{Conclusion}
In this paper, we addressed the challenge of simulating human survey responses by introducing \textbf{MARK (Multi-stAge Reasoning frameworK)}, a novel framework grounded in MBTI personality theory. Through a multi-stage approach that considers stress levels, personality types, and their associated cognitive functions, MARK provides a more psychologically grounded method for predicting survey responses.

Our experiments on the World Values Survey demonstrate that MARK can outperform existing baselines by 10\%. This improvement suggests that modeling personality-based cognitive processes offers a promising direction for enhancing survey simulation accuracy. Furthermore, our analysis reveals that our group-level experiment, weighted cognitive imitation, and stress-informed reasoning reached robustness on uncertain personalities.

Moving forward, this work opens new avenues for leveraging AI technologies to create low-divergence, personalized survey simulations, potentially reducing the reliance on costly and time-consuming human experiments in social survey research. 
Future work could focus on improving personality prediction accuracy and expanding the framework to handle a broader range of survey types and domains.

\section*{Limitations} 
\paragraph{Accurate subject personality}  
Our study examines the personality traits of human subjects in the World Values Survey through predictions made by LLMs based on human and personality preferences, presenting a complex challenge. Our findings suggest that incorporating human personalities can enhance simulation performance; however, this may also restrict the applicability of the findings without real-world personality information. We hope that future survey investigations will consider incorporating personality as a key human feature to provide more precise psychological insights.

\paragraph{Language and countries coverage} 
Our study employs English and Chinese prompts for simulations conducted in the United States and China, highlighting the distinct cultural differences related to cognition, language, perception, and reasoning \citep{Cao22US}. While the diverse populations of both the U.S. and China illustrate significant East-West cultural contrasts, this approach does not fully capture the broader cultural diversity present worldwide. To address these limitations, future research should consider incorporating additional countries and languages to strengthen the robustness of our framework.

\paragraph{Model choices}
Due to considerations regarding computational resources, we conducted experiments with closed-source models through API, selecting a limited number of models for validation. We also implemented our framework on open-source models such as Llama-3.1-7B/13B. Experimental results showed they are limited in following complex instructions and conducting cognitive simulations, likely due to the length and complexity of system instructions and human features. In our future work, we plan to explore a variety of open-source models that feature large context windows and robust instruction-following capabilities. This approach will enable us to uncover valuable insights. We believe the conclusions we draw will significantly contribute to future research, encouraging further exploration of additional models to enhance our understanding of personalities in simulations.

\paragraph{Psychological Validity vs. Computational Tractability} We acknowledge the well-documented empirical limitations of the Myers-Briggs Type Indicator (MBTI) framework in psychological research, including reduced reliability~\citep{boyle95myers1995} and validity compared to the Big Five~\citep{Graziotin21Psychometrics}. However, our methodological choice prioritizes computational tractability, as MBTI's \textbf{hierarchical function stack} provides explicit precedence rules for information processing, \textbf{systematic cognitive load balancing}~\cite{Maria20Computational}, and \textbf{adaptive reasoning capabilities}~\cite{Rogers89Computational}. While the Big Five excels at empirical personality measurement~\cite{Colin15Cybernetic}, it lacks the hierarchical cognitive process models essential for structured multi-step reasoning, as it \textbf{identifies what cognitive differences} exist but provides \textbf{limited frameworks for how these differences operate} as integrated systems. Future work should explore hierarchical frameworks grounded in empirically validated personality dimensions.

\section*{Ethics Statement}
This research upholds strict ethical standards, ensuring that all datasets, large language models, and prompt configurations utilized are sourced from open-access repositories and are appropriately licensed by their original creators.

Through this work, we aim to provide valuable tools for social survey simulations and investigations while promoting social equity and sustainability. We are dedicated to fostering an understanding and appreciation of cultural diversity through AI technologies, emphasizing the importance of employing responsible approaches when simulating human preferences that prevent misunderstandings or biases against any cultural group. Additionally, we are committed to maintaining transparency throughout our research process and actively engaging with stakeholders to ensure that our efforts align with the highest ethical standards.

We acknowledge concerns about demographic stereotyping and clarify that our framework views demographics as \textbf{indirect signals} of environmental pressures, not fixed predictors of personality traits. Unlike methods that directly link demographics to opinions, our approach models personality as a latent construct influenced by contextual factors. While this reduces oversimplification compared to direct mappings, we recognize the need for future work to include additional mediating variables to further address bias risks.

\section*{Acknowledgements}
We thank the anonymous reviewers for their valuable feedback and constructive comments that helped improve this work. We are grateful to Yuchu Tian and Caicai Guo for their insightful discussions and suggestions.

This work was supported by the National Key Research and Development Program of China under Grants 2022YFC3300801, the Research Project of China Publishing Promotion Association under Grants 2025ZBCH-JYYB17, and the Natural Science Foundation of Hubei Province (CN) under Grant No. 2025AFB078.

\bibliography{anthology, custom}

\begin{thebibliography}{40}
\providecommand{\natexlab}[1]{#1}

\bibitem[{Aher et~al.(2023)Aher, Arriaga, and Kalai}]{aher2023using}
Gati~V Aher, Rosa~I Arriaga, and Adam~Tauman Kalai. 2023.
\newblock Using large language models to simulate multiple humans and replicate human subject studies.
\newblock In \emph{International Conference on Machine Learning}, pages 337--371. PMLR.

\bibitem[{AlKhamissi et~al.(2024)AlKhamissi, ElNokrashy, Alkhamissi, and Diab}]{alkhamissi-etal-2024-investigating}
Badr AlKhamissi, Muhammad ElNokrashy, Mai Alkhamissi, and Mona Diab. 2024.
\newblock \href {https://doi.org/10.18653/v1/2024.acl-long.671} {Investigating cultural alignment of large language models}.
\newblock In \emph{Proceedings of the 62nd Annual Meeting of the Association for Computational Linguistics (Volume 1: Long Papers)}, pages 12404--12422, Bangkok, Thailand. Association for Computational Linguistics.

\bibitem[{Argyle et~al.(2023)Argyle, Busby, Fulda, Gubler, Rytting, and Wingate}]{argyle2023out}
Lisa~P Argyle, Ethan~C Busby, Nancy Fulda, Joshua~R Gubler, Christopher Rytting, and David Wingate. 2023.
\newblock Out of one, many: Using language models to simulate human samples.
\newblock \emph{Political Analysis}, 31(3):337--351.

\bibitem[{Arora et~al.(2023)Arora, Kaffee, and Augenstein}]{arora-etal-2023-probing}
Arnav Arora, Lucie-aim{\'e}e Kaffee, and Isabelle Augenstein. 2023.
\newblock \href {https://doi.org/10.18653/v1/2023.c3nlp-1.12} {Probing pre-trained language models for cross-cultural differences in values}.
\newblock In \emph{Proceedings of the First Workshop on Cross-Cultural Considerations in NLP (C3NLP)}, pages 114--130, Dubrovnik, Croatia. Association for Computational Linguistics.

\bibitem[{Badshah and Sajjad(2024)}]{badshah24reference}
Sher Badshah and Hassan Sajjad. 2024.
\newblock \href {https://doi.org/10.48550/arXiv.2408.09235} {Reference-guided verdict: Llms-as-judges in automatic evaluation of free-form text}.
\newblock \emph{arXiv e-prints}, arXiv:2408.09235.

\bibitem[{Benkler et~al.(2023)Benkler, Mosaphir, Friedman, Smart, and Schmer-Galunder}]{benkler2023assessing}
Noam Benkler, Drisana Mosaphir, Scott Friedman, Andrew Smart, and Sonja Schmer-Galunder. 2023.
\newblock Assessing {LLM}s for moral value pluralism.
\newblock \emph{arXiv preprint arXiv:2312.10075}.

\bibitem[{Bisbee et~al.(2023)Bisbee, Clinton, Dorff, Kenkel, and Larson}]{bisbee2023synthetic}
James Bisbee, Joshua~D Clinton, Cassy Dorff, Brenton Kenkel, and Jennifer~M Larson. 2023.
\newblock Synthetic replacements for human survey data? the perils of large language models.
\newblock \emph{Political Analysis}, pages 1--16.

\bibitem[{Boyle()}]{boyle95myers1995}
Gregory~J. Boyle.
\newblock \href {https://doi.org/10.1111/j.1742-9544.1995.tb01750.x} {Myers-briggs type indicator ({MBTI}): Some psychometric limitations}.
\newblock 30(1):71--74.

\bibitem[{Cao et~al.(2024)Cao, Carstensen, Gao, and Frank}]{Cao22US}
Anjie Cao, Alexandra Carstensen, Shan Gao, and Michael~C Frank. 2024.
\newblock United {States-China} differences in cognition and perception across 12 tasks: Replicability, robustness, and within-culture variation.
\newblock \emph{J Exp Psychol Gen}.

\bibitem[{Cao et~al.(2025)Cao, Liu, Arora, Augenstein, R{\"o}ttger, and Hershcovich}]{cao-etal-2025-specializing}
Yong Cao, Haijiang Liu, Arnav Arora, Isabelle Augenstein, Paul R{\"o}ttger, and Daniel Hershcovich. 2025.
\newblock \href {https://aclanthology.org/2025.naacl-long.162/} {Specializing large language models to simulate survey response distributions for global populations}.
\newblock In \emph{Proceedings of the 2025 Conference of the Nations of the Americas Chapter of the Association for Computational Linguistics: Human Language Technologies (Volume 1: Long Papers)}, pages 3141--3154, Albuquerque, New Mexico. Association for Computational Linguistics.

\bibitem[{Cao et~al.(2023)Cao, Zhou, Lee, Cabello, Chen, and Hershcovich}]{cao-etal-2023-assessing}
Yong Cao, Li~Zhou, Seolhwa Lee, Laura Cabello, Min Chen, and Daniel Hershcovich. 2023.
\newblock \href {https://doi.org/10.18653/v1/2023.c3nlp-1.7} {Assessing cross-cultural alignment between {C}hat{GPT} and human societies: An empirical study}.
\newblock In \emph{Proceedings of the First Workshop on Cross-Cultural Considerations in NLP (C3NLP)}, pages 53--67, Dubrovnik, Croatia. Association for Computational Linguistics.

\bibitem[{Cheng et~al.(2023)Cheng, Durmus, and Jurafsky}]{cheng-etal-2023-marked}
Myra Cheng, Esin Durmus, and Dan Jurafsky. 2023.
\newblock \href {https://doi.org/10.18653/v1/2023.acl-long.84} {Marked personas: Using natural language prompts to measure stereotypes in language models}.
\newblock In \emph{Proceedings of the 61st Annual Meeting of the Association for Computational Linguistics (Volume 1: Long Papers)}, pages 1504--1532, Toronto, Canada. Association for Computational Linguistics.

\bibitem[{Clark(2002)}]{Clark02stress}
Phoebe Clark. 2002.
\newblock Stress and type dynamics.
\newblock \emph{Bulletin of Psychological Type}, 25.3:41--43.

\bibitem[{Clawson(2008)}]{Clawson08Myers}
James Clawson. 2008.
\newblock Myers-briggs type indicator.

\bibitem[{Deshpande et~al.(2023)Deshpande, Murahari, Rajpurohit, Kalyan, and Narasimhan}]{deshpande-etal-2023-toxicity}
Ameet Deshpande, Vishvak Murahari, Tanmay Rajpurohit, Ashwin Kalyan, and Karthik Narasimhan. 2023.
\newblock \href {https://doi.org/10.18653/v1/2023.findings-emnlp.88} {Toxicity in chatgpt: Analyzing persona-assigned language models}.
\newblock In \emph{Findings of the Association for Computational Linguistics: EMNLP 2023}, pages 1236--1270, Singapore. Association for Computational Linguistics.

\bibitem[{DeYoung(2015)}]{Colin15Cybernetic}
Colin~G. DeYoung. 2015.
\newblock \href {https://doi.org/10.1016/j.jrp.2014.07.004} {Cybernetic big five theory}.
\newblock \emph{Journal of Research in Personality}, 56:33--58.
\newblock Integrative Theories of Personality.

\bibitem[{Durmus et~al.(2024)Durmus, Nguyen, Liao, Schiefer, Askell, Bakhtin, Chen, Hatfield-Dodds, Hernandez, Joseph, Lovitt, McCandlish, Sikder, Tamkin, Thamkul, Kaplan, Clark, and Ganguli}]{durmus2024measuringrepresentationsubjectiveglobal}
Esin Durmus, Karina Nguyen, Thomas~I. Liao, Nicholas Schiefer, Amanda Askell, Anton Bakhtin, Carol Chen, Zac Hatfield-Dodds, Danny Hernandez, Nicholas Joseph, Liane Lovitt, Sam McCandlish, Orowa Sikder, Alex Tamkin, Janel Thamkul, Jared Kaplan, Jack Clark, and Deep Ganguli. 2024.
\newblock \href {https://arxiv.org/abs/2306.16388} {Towards measuring the representation of subjective global opinions in language models}.

\bibitem[{Eckstein and Collins(2020)}]{Maria20Computational}
Maria~K. Eckstein and Anne G.~E. Collins. 2020.
\newblock \href {https://doi.org/10.1073/pnas.1912330117} {Computational evidence for hierarchically structured reinforcement learning in humans}.
\newblock \emph{Proceedings of the National Academy of Sciences}, 117(47):29381--29389.

\bibitem[{Fang et~al.(2020)Fang, Chung, and Wang}]{fang20impact}
Siqi Fang, Man~Cheung Chung, and Yabing Wang. 2020.
\newblock \href {https://doi.org/10.3389/fpsyg.2020.00992} {The impact of past trauma on psychological distress: The roles of defense mechanisms and alexithymia}.
\newblock \emph{Frontiers in Psychology}, 11.

\bibitem[{GLM et~al.(2024)GLM, :, Zeng, Xu, Wang, Zhang, Yin, Zhang, Rojas, Feng, Zhao, Lai, Yu, Wang, Sun, Zhang, Cheng, Gui, Tang, Zhang, Sun, Li, Zhao, Wu, Zhong, Liu, Huang, Zhang, Zheng, Lu, Duan, Zhang, Cao, Yang, Tam, Zhao, Liu, Xia, Zhang, Gu, Lv, Liu, Liu, Yang, Song, Zhang, An, Xu, Niu, Yang, Li, Bai, Dong, Qi, Wang, Yang, Du, Hou, and Wang}]{glm2024chatglmfamilylargelanguage}
Team GLM, :, Aohan Zeng, Bin Xu, Bowen Wang, Chenhui Zhang, Da~Yin, Dan Zhang, Diego Rojas, Guanyu Feng, Hanlin Zhao, Hanyu Lai, Hao Yu, Hongning Wang, Jiadai Sun, Jiajie Zhang, Jiale Cheng, Jiayi Gui, Jie Tang, Jing Zhang, Jingyu Sun, Juanzi Li, Lei Zhao, Lindong Wu, Lucen Zhong, Mingdao Liu, Minlie Huang, Peng Zhang, Qinkai Zheng, Rui Lu, Shuaiqi Duan, Shudan Zhang, Shulin Cao, Shuxun Yang, Weng~Lam Tam, Wenyi Zhao, Xiao Liu, Xiao Xia, Xiaohan Zhang, Xiaotao Gu, Xin Lv, Xinghan Liu, Xinyi Liu, Xinyue Yang, Xixuan Song, Xunkai Zhang, Yifan An, Yifan Xu, Yilin Niu, Yuantao Yang, Yueyan Li, Yushi Bai, Yuxiao Dong, Zehan Qi, Zhaoyu Wang, Zhen Yang, Zhengxiao Du, Zhenyu Hou, and Zihan Wang. 2024.
\newblock \href {https://arxiv.org/abs/2406.12793} {Chatglm: A family of large language models from glm-130b to glm-4 all tools}.
\newblock \emph{Preprint}, arXiv:2406.12793.

\bibitem[{Graziotin et~al.(2021)Graziotin, Lenberg, Feldt, and Wagner}]{Graziotin21Psychometrics}
Daniel Graziotin, Per Lenberg, Robert Feldt, and Stefan Wagner. 2021.
\newblock \href {https://doi.org/10.1145/3469888} {Psychometrics in behavioral software engineering: A methodological introduction with guidelines}.
\newblock \emph{ACM Trans. Softw. Eng. Methodol.}, 31(1).

\bibitem[{Hall(1989)}]{Rogers89Computational}
Rogers~P. Hall. 1989.
\newblock \href {https://doi.org/10.1016/0004-3702(89)90003-9} {Computational approaches to analogical reasoning: A comparative analysis}.
\newblock \emph{Artificial Intelligence}, 39(1):39--120.

\bibitem[{Horton(2023)}]{horton2023large}
John~J Horton. 2023.
\newblock Large language models as simulated economic agents: What can we learn from homo silicus?
\newblock Technical report, National Bureau of Economic Research.

\bibitem[{Hwang et~al.(2023)Hwang, Majumder, and Tandon}]{hwang-etal-2023-aligning}
EunJeong Hwang, Bodhisattwa Majumder, and Niket Tandon. 2023.
\newblock \href {https://doi.org/10.18653/v1/2023.findings-emnlp.393} {Aligning language models to user opinions}.
\newblock In \emph{Findings of the Association for Computational Linguistics: EMNLP 2023}, pages 5906--5919, Singapore. Association for Computational Linguistics.

\bibitem[{Kwok et~al.(2024)Kwok, Bravansky, and Griffin}]{kwok2024evaluating}
Louis Kwok, Michal Bravansky, and Lewis~D Griffin. 2024.
\newblock Evaluating cultural adaptability of a large language model via simulation of synthetic personas.
\newblock \emph{arXiv preprint arXiv:2408.06929}.

\bibitem[{Li et~al.(2024)Li, Jiang, Hwang, Kim, Santy, Sorensen, Lin, Dziri, Ren, and Choi}]{li2024culturegenrevealingglobalcultural}
Huihan Li, Liwei Jiang, Jena~D. Hwang, Hyunwoo Kim, Sebastin Santy, Taylor Sorensen, Bill~Yuchen Lin, Nouha Dziri, Xiang Ren, and Yejin Choi. 2024.
\newblock \href {https://arxiv.org/abs/2404.10199} {Culture-gen: Revealing global cultural perception in language models through natural language prompting}.
\newblock \emph{Preprint}, arXiv:2404.10199.

\bibitem[{Liu et~al.(2025)Liu, Cao, Wu, Qiu, Gu, Liu, and Hershcovich}]{LIU25Towards}
Haijiang Liu, Yong Cao, Xun Wu, Chen Qiu, Jinguang Gu, Maofu Liu, and Daniel Hershcovich. 2025.
\newblock \href {https://doi.org/10.1016/j.ipm.2025.104099} {Towards realistic evaluation of cultural value alignment in large language models: Diversity enhancement for survey response simulation}.
\newblock \emph{Information Processing \& Management}, 62(4):104099.

\bibitem[{Luo et~al.(2025)Luo, Liu, Zhang, Gao, and Gu}]{Luo2025ASO}
Yitian Luo, Yu~Liu, Lu~Zhang, Feng Gao, and Jinguang Gu. 2025.
\newblock \href {https://api.semanticscholar.org/CorpusID:277681369} {A survey on quality evaluation of instruction fine-tuning datasets for large language models}.
\newblock \emph{Data Intell.}, 7:527--566.

\bibitem[{Mitchell(2001)}]{Mitchell01type}
Wayne~D. Mitchell. 2001.
\newblock A full dynamic model of type.
\newblock \emph{Journal of Psychological Type}, pages 12--28.

\bibitem[{Mou et~al.(2024)Mou, Ding, He, Wang, Liang, Zhang, Sun, Lin, Zhou, Huang, and Wei}]{mou2024individualsocietysurveysocial}
Xinyi Mou, Xuanwen Ding, Qi~He, Liang Wang, Jingcong Liang, Xinnong Zhang, Libo Sun, Jiayu Lin, Jie Zhou, Xuanjing Huang, and Zhongyu Wei. 2024.
\newblock \href {https://arxiv.org/abs/2412.03563} {From individual to society: A survey on social simulation driven by large language model-based agents}.
\newblock \emph{Preprint}, arXiv:2412.03563.

\bibitem[{Myers(1998)}]{myers1998mbti}
I.B. Myers. 1998.
\newblock \href {https://books.google.com.hk/books?id=XPbRQwAACAAJ} {\emph{MBTI Manual: A Guide to the Development and Use of the Myers-Briggs Type Indicator}}.
\newblock Consulting Psychologists Press.

\bibitem[{Oncevay et~al.(2020)Oncevay, Haddow, and Birch}]{oncevay-etal-2020-bridging}
Arturo Oncevay, Barry Haddow, and Alexandra Birch. 2020.
\newblock \href {https://doi.org/10.18653/v1/2020.emnlp-main.187} {Bridging linguistic typology and multilingual machine translation with multi-view language representations}.
\newblock In \emph{Proceedings of the 2020 Conference on Empirical Methods in Natural Language Processing (EMNLP)}, pages 2391--2406, Online. Association for Computational Linguistics.

\bibitem[{{OpenAI} et~al.(2024){OpenAI}, {:}, {Hurst}, {Lerer}, and {Goucher}}]{gpt-4o}
{OpenAI}, {:}, Aaron {Hurst}, Adam {Lerer}, and Adam P. et~al. {Goucher}. 2024.
\newblock \href {https://doi.org/10.48550/arXiv.2410.21276} {{GPT-4o System Card}}.
\newblock \emph{arXiv e-prints}, arXiv:2410.21276.

\bibitem[{Rentfrow et~al.(2013)Rentfrow, Gosling, Jokela, Stillwell, Kosinski, and Potter}]{Rentfrow2013Divided}
Peter~J. Rentfrow, Samuel~D. Gosling, Markus Jokela, David~J. Stillwell, Michal Kosinski, and Jeff Potter. 2013.
\newblock Divided we stand: Three psychological regions of the {United} {States} and their political, economic, social, and health correlates.
\newblock \emph{Journal of Personality and Social Psychology}, 105(6):996--1012.

\bibitem[{Rubner et~al.(1998)Rubner, Tomasi, and Guibas}]{rubner1998metric}
Yossi Rubner, Carlo Tomasi, and Leonidas~J Guibas. 1998.
\newblock A metric for distributions with applications to image databases.
\newblock In \emph{Sixth international conference on computer vision (IEEE Cat. No. 98CH36271)}, pages 59--66. IEEE.

\bibitem[{Santurkar et~al.(2023)Santurkar, Durmus, Ladhak, Lee, Liang, and Hashimoto}]{santurkar2023opinionslanguagemodelsreflect}
Shibani Santurkar, Esin Durmus, Faisal Ladhak, Cinoo Lee, Percy Liang, and Tatsunori Hashimoto. 2023.
\newblock \href {https://arxiv.org/abs/2303.17548} {Whose opinions do language models reflect?}
\newblock \emph{Preprint}, arXiv:2303.17548.

\bibitem[{Shen et~al.(2024)Shen, Logeswaran, Lee, Lee, Poria, and Mihalcea}]{shen-etal-2024-understanding}
Siqi Shen, Lajanugen Logeswaran, Moontae Lee, Honglak Lee, Soujanya Poria, and Rada Mihalcea. 2024.
\newblock \href {https://doi.org/10.18653/v1/2024.naacl-long.316} {Understanding the capabilities and limitations of large language models for cultural commonsense}.
\newblock In \emph{Proceedings of the 2024 Conference of the North American Chapter of the Association for Computational Linguistics: Human Language Technologies (Volume 1: Long Papers)}, pages 5668--5680, Mexico City, Mexico. Association for Computational Linguistics.

\bibitem[{Zhao et~al.(2024{\natexlab{a}})Zhao, Mondal, Tandon, Dillion, Gray, and Gu}]{zhao-etal-2024-worldvaluesbench}
Wenlong Zhao, Debanjan Mondal, Niket Tandon, Danica Dillion, Kurt Gray, and Yuling Gu. 2024{\natexlab{a}}.
\newblock \href {https://aclanthology.org/2024.lrec-main.1539/} {{W}orld{V}alues{B}ench: A large-scale benchmark dataset for multi-cultural value awareness of language models}.
\newblock In \emph{Proceedings of the 2024 Joint International Conference on Computational Linguistics, Language Resources and Evaluation (LREC-COLING 2024)}, pages 17696--17706, Torino, Italia. ELRA and ICCL.

\bibitem[{Zhao et~al.(2024{\natexlab{b}})Zhao, Mondal, Tandon, Dillion, Gray, and Gu}]{zhao2024world}
Wenlong Zhao, Debanjan Mondal, Niket Tandon, Danica Dillion, Kurt Gray, and Yuling Gu. 2024{\natexlab{b}}.
\newblock Worldvaluesbench: A large-scale benchmark dataset for multi-cultural value awareness of language models.
\newblock In \emph{Proceedings of the 2024 Joint International Conference on Computational Linguistics, Language Resources and Evaluation (LREC-COLING 2024)}, pages 17696--17706.

\bibitem[{Zhou et~al.(2023)Zhou, Karamolegkou, Chen, and Hershcovich}]{zhou-etal-2023-cultural}
Li~Zhou, Antonia Karamolegkou, Wenyu Chen, and Daniel Hershcovich. 2023.
\newblock \href {https://doi.org/10.18653/v1/2023.findings-emnlp.845} {Cultural compass: Predicting transfer learning success in offensive language detection with cultural features}.
\newblock In \emph{Findings of the Association for Computational Linguistics: EMNLP 2023}, pages 12684--12702, Singapore. Association for Computational Linguistics.

\end{thebibliography}

\appendix
\section{Cognition Functions}
\label{app:cfun}
The cognition processes of the MBTI personality type dynamics can be formulated into eight functions:
\begin{enumerate}
    \item Extroverted Sensing (Se): Acts on concrete data in the here and now. Likes to experience the world—active, talkative, and social. Trusts the present, what is tangible and real. When the stress level is high, this function will tend to be overindulgent, hyperactive, and overly talkative.
    \item Introverted Sensing (Si): Compares present facts and situations to past experience. Excellent recall for specific details. Trusts and remembers the past. Stores sensory data that is important to them for future use. When the stress level is high, this function will tend to be dogmatic, obsess about unimportant data, and withdraw.
    \item Extroverted Intuition (Ne): Sees possibilities in the external world. Enthusiastic and enjoys networking. Trusts the big picture, and forms patterns and connections, which can then be shared with others. When the stress level is high, this function will tend to be over the top, swamped with options, and change for the sake of change.
    \item Introverted Intuition (Ni): Can appear visionary. Connects unconscious images, themes, and connections to see things in new ways. Brainstorm internally with themselves. Trusts and relies on inner insights, which may be hard for others to understand. When the stress level is high, this function will tend to have unrealistic visions, only accept data that supports their theories, and make things overcomplicated.
    \item Extroverted Thinking (Te): Seeks logic and consistency in the outside world. Concern for external laws and rules. Logical, analytical decision-makers who organize the environment to achieve goals. When the stress level is high, this function will tend to be detached, cold, overly rational, and critique the lack of logic in others.
    \item Introverted Thinking (Ti): Seeks internal consistency and logic of ideas. Trust's internal framework, which may be difficult to explain to others. Experience a depth of concentration that is objective and analytical. When the stress level is high, this function will tend to be an obsessive search for the truth, detached, look only at the cons, driven like a machine out of control.
    \item Extroverted Feeling (Fe): Seeks harmony with and between people in the outside world. Interpersonal and cultural values are important. Encouraging and interested in others. When the stress level is high, this function will tend to be insistent, meaning that they know what is best for everyone, are intrusive, ignore problems, and force superficial harmony.
    \item Introverted Feeling (Fi): Seeks harmony of action and thoughts with personal values. May not always articulate those values. Empathetic, sensitive, and idealistic. When stress levels are high, this function will tend to carry the weight of the world on their shoulders, be hypersensitive, pompous, and feel sorry for themselves.
\end{enumerate}

\section{Personality Selection Algorithm}
\label{sec:person_predict}
In this section, we present the personality selection algorithm during Stage 2 of the MARK framework in algorithm \ref{alg1}.

\section{Reasoning Prompts}
\label{app:prompts}
In this section, we present the system prompts designed for each stage within the MARK system.

For the \textbf{human stress analysis stage}, as presented in Table \ref{tbl:fea_ana_prompt}, we first instruct the model to assign a stress level based on human features. Subsequently, the model automatically generates sociodemographic prompts by filtering out insignificant features.

For the \textbf{personality prediction}, as shown in Table \ref{tbl:person_pred_prompt}, we employ a similar system prompt to guide the model in predicting the dominant and auxiliary processes based on human features. To ensure the quality of process combinations, we derive the tertiary and inferior processes using MBTI-type dynamics theory.

During the survey response generation phase, we employ the \textbf{Cognitive Reasoning Stage}, guided by the prompts in Table \ref{tbl:cog_prompt}, to produce simulation results based on the predicted personality processes. First, the model evaluates the situational stress impact on each of the four processes using sociodemographic prompts and the given survey question. This step determines whether the stress level is sufficiently high to trigger the negative effects of a process. Next, the model generates distinct descriptions for each process and produces potential responses for each one.

After generating answers for the four processes, the \textbf{Synthesis Stage}, prompted as shown in Table \ref{tbl:syn_prompt}, synthesizes the results based on the importance weights provided by the reasoning stage. Finally, it consolidates the outputs to determine the simulated preference.

\section{Data Construction Details}
In this section, we present the implementation details for constructing the experiment dataset.

\subsection{Group-level Demographic Clustering}
\label{sec:cls_ana}
\begin{figure}
    \centering
    \includegraphics[width=0.9\linewidth]{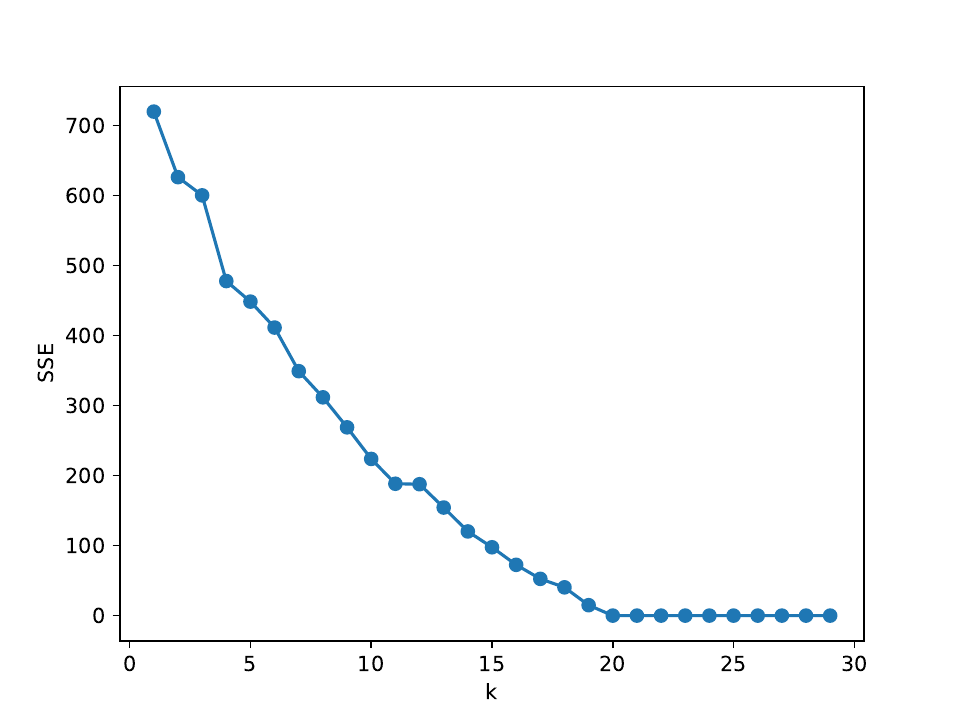}
    % \label{fig:cha_cluster-k}
    \caption{The Silhouette Analysis for demographic clustering to determine the optimal number of clusters using KMeans clustering. The optimal number for demographic clustering is 20.}
    \label{fig:cha_cluster-k}
\end{figure}
The analysis in Figure \ref{fig:cha_cluster-k} revealed an optimal configuration of 20 clusters, and we gathered subjects according to these clustering results.

\subsection{Personality Augmentation for Dataset}
\label{sec:data_aug} 
We instructed the LLMs following Table \ref{tbl:aug_prompt} to predict the likelihood of each role and personality type selecting a specific answer based on a given question. For individual data, we identified the role with the highest probability and then determined the most probable personality within that role. This process yielded a comprehensive personality profile for each survey subject.

\subsection{Full Experiment Results of feature Sensitivity}
As presented in Table \ref{tbl:full_result}, our method MARK obtained the highest simulation accuracy on most clusters for both feature clustering. 

For distribution divergence, in Table \ref{tbl:full_jsd} and \ref{tbl:full_emd}, MARK presents lower performance than Demo.+Ideo+Opiniontop3; this might be caused by a potential preference leakage during the inference time, where we provide the top relevant answers from the subject gold labels. 

We also provide the detailed results of the different personality strategies for both feature clustering in Table \ref{tbl:full_persona_chara}. With the personalities assumed from the subjects' preferences provided, the simulation performance soared on both clustering settings both improving accuracy and reducing divergence.

We provide prediction accuracy for feature clustering in Table \ref{tbl:persona_acc}. The result suggests that the model has different sensitivities to predicting personalities based on similar human demographic information. Generally, it achieved high accuracy in predicting high-level personalities (role), while having difficulties predicting the correct low-level types.

\section{Reasoning Demo}
To better understand the model's behaviors during survey completion, we present a Reasoning demo (model prediction in {\sffamily sans serif} or tables) that shows how the reasoning works in practice.

\paragraph{Background} Here we present some background information for this case:

\begin{itemize}
    \item \textbf{Case Group}: This group of respondents, predominantly female and middle-aged, shares similar demographic and socioeconomic characteristics.
    \item \textbf{Subject Profile}: The unique profile of the group representative (selected by the k-means algorithm) is presented in Table \ref{tab:demoinfo}.
    \item \textbf{Survey Question}: For family, would you say it is very important, rather important, not very important, or not important at all?
    \item \textbf{Options}: (A) Very important, (B) Rather important, (C) Not very important, (D) Not important at all
\end{itemize}

\begin{table}[]
\resizebox{\linewidth}{!}{%
\begin{tabular}{ll}
\toprule
\textbf{Category} & \textbf{Description}                            \\ \midrule
Location          & Urban, Georgia, USA                             \\
Education         & Upper secondary education (high school diploma) \\
Marital Status    & Married                                         \\
Family            & 2 children                                      \\
Age               & 44 years old                                    \\
Occupation        & Homemaker, not employed outside the home        \\
Citizenship       & US citizen                                      \\
Religion          & Catholic                                        \\
Ethnicity         & US, two or more, non-Hispanic                   \\ \bottomrule
\end{tabular}%
}
\caption{This table summarizes the key characteristics of the group representative, providing a comprehensive overview of her demographic, socioeconomic, and cultural background.}
\label{tab:demoinfo}
\end{table}

\begin{table*}[]
\resizebox{\textwidth}{!}{%
\begin{tabular}{lllll}
\toprule
\textbf{Demographic Feature} &
  \textbf{Value} &
  \textbf{\begin{tabular}[c]{@{}l@{}}Stress \\ Level\end{tabular}} &
  \textbf{\begin{tabular}[c]{@{}l@{}}Stress \\ Category\end{tabular}} &
  \textbf{Rationale} \\ \midrule
Financial Behavior &
  \begin{tabular}[c]{@{}l@{}}Spent savings and borrowed \\ money\end{tabular} &
  70/100 &
  Economic &
  \begin{tabular}[c]{@{}l@{}}Negative financial trajectory indicates severe resource \\ pressure affecting security priorities\end{tabular} \\
Employment Status &
  \begin{tabular}[c]{@{}l@{}}Homemaker not otherwise \\ employed\end{tabular} &
  40/100 &
  Career/Identity &
  \begin{tabular}[c]{@{}l@{}}Lack of formal employment creates moderate stress \\ around financial independence and social identity\end{tabular} \\
Age &
  44 years &
  25/100 &
  Life Stage &
  \begin{tabular}[c]{@{}l@{}}Middle-age responsibilities bring moderate complexity \\ to decision-making processes\end{tabular} \\ 

... & ... & ... & ... & ... \\ \bottomrule
\end{tabular}%
}
\caption{Demographic Profile and Stress Analysis for Framework Demonstration}
\label{tab:stress_ana_demo}
\end{table*}

\begin{table}[]
\resizebox{\linewidth}{!}{%
\begin{tabular}{llll}
\toprule
\textbf{\begin{tabular}[c]{@{}l@{}}Reasoning \\ Stage\end{tabular}} &
  \textbf{\begin{tabular}[c]{@{}l@{}}Cognitive \\ Process\end{tabular}} &
  \textbf{\begin{tabular}[c]{@{}l@{}}Stress \\ Impact\end{tabular}} &
  \textbf{Impact Rationale} \\ \midrule
Dominant &
  \begin{tabular}[c]{@{}l@{}}Introverted \\ Sensing (Si)\end{tabular} &
  Positive &
  \begin{tabular}[c]{@{}l@{}}Past security experiences guide current \\ decision-making effectively\end{tabular} \\
Auxiliary &
  \begin{tabular}[c]{@{}l@{}}Extroverted \\ Feeling (Fe)\end{tabular} &
  Positive &
  \begin{tabular}[c]{@{}l@{}}Community values support importance \\ of family security\end{tabular} \\
Tertiary &
  \begin{tabular}[c]{@{}l@{}}Introverted \\ Thinking (Ti)\end{tabular} &
  Positive &
  \begin{tabular}[c]{@{}l@{}}Logical analysis remains functional \\ despite stress\end{tabular} \\
Inferior &
  \begin{tabular}[c]{@{}l@{}}Extraverted \\ Intuition (Ne)\end{tabular} &
  Positive &
  \begin{tabular}[c]{@{}l@{}}Stress level manageable, allows \\ consideration of possibilities\end{tabular} \\ \bottomrule
\end{tabular}%
}
\caption{Cognitive Process Stress Impact Analysis}
\label{tab:cog_impact}
\end{table}

\paragraph{Stage 1: Stress Analysis - Sociodemographic Context Generation} GPT-4o analyzed stress level are in Table \ref{tab:stress_ana_demo}. 
The generated sociodemographic context prompt for features with minimal impact on the simulation is:

{\small
The dropped characters represent a demographic situated in urban Georgia, navigating the moderate pressures of middle age as a native citizen. }

{\small The household is small, comprising two adults without direct parental support, reflecting a stable yet limited economic situation. The educational background is primarily secondary, indicating potential limits in career advancement opportunities. Being married provides some relational stability, while their involvement in religious activities offers additional community support. 
}

\paragraph{Stage 2: Personality Prediction} Based on the analyzed demographic features and sociodemographic context prompt, GPT-4o predicted the \sethlcolor{lightblue}\hl{ISFJ} as her personality and provided the following cognition process:

{\small \textbf{Introverted Sensing (Si)}: Strong preference for proven, secure approaches}

{\small \textbf{Extraverted Feeling (Fe)}: Concern for family/community welfare}

{\small \textbf{Introverted Thinking (Ti)}: Practical evaluation of options}

{\small \textbf{Extraverted Intuition (Ne)}: Limited exploration due to risk aversion}

\paragraph{Stage 3: Cognitive Reasoning} First, the model evaluates how survey stress, combining social stress (from stage 1), affects each cognitive process. Analyzed results are in Table \ref{tab:cog_impact}. All processes remain functional (positive impact) as the stress level (46.5/100) does not reach the threshold for cognitive disruption.

Then the model reason over the function description and choose an option as an answer for the survey question. In this case, GPT-4o predicted:

{\small Dominate process (Si): (A) Very important, weight - 0.6}

{\small Auxiliary process (Fe): (A) Very important, weight - 0.5}

{\small Tertiary process (Ti): (C) Not very important, weight - 0.3}

{\small Inferior process (Ne): (D) Not at all important, weight - 0.2}

\paragraph{Stage 4: Synthesis} Finally, GPT-4o reviewed previous predictions and combined the results according to the reviewed weight and chose one option as the final answer:

{\small
\textbf{Conclusion}: (A) Very important}

{\small \textbf{Reasoning Summary}: The final conclusion of "Very important" emerges from the dominant Introverted Sensing process (60\% weight) and auxiliary Extroverted Feeling process (50\% weight), both strongly aligning with cultural and community emphasis on family. Although the tertiary and inferior processes suggested "Not very important" and "Not at all important" view, their lesser weights (30\% and 20\%, respectively) resulted in a decisive emphasis on the highest importance of family.
}

\begin{algorithm*}
    \SetKwFunction{getdesc}{get\_all\_process\_desc}
    \SetKwFunction{gatherinput}{gather\_input}
    \SetKwFunction{gen}{generate}
    \SetKwFunction{can}{get\_next\_process}
    \SetKwFunction{dfs}{dfs}
    
    \caption{Personality process selection algorithm for subject $j$.}
    \label{alg1}
        \KwIn{Selection system prompt $P_{\text{sys}}$, generated sociodemographic prompting $P_{\text{demo}}$, overall stress $S_{\text{overall}_j}$, }
        \KwOut{Selected four cognition processes $p_{\text{cog}_j}$ and corresponding personality type $T_j$}
        \nonl \textbf{Hyperparameters}:
        \begin{enumerate}
            \item Model generation temperature $t$
            \item Model generation max tokens $m_{\text{token}}$
        \end{enumerate}

        \textbf{Functions}: 
        \begin{enumerate}
            \item Access all cognition process descriptions: \getdesc{}
            \item Access process candidates with description: \can{process name, selected process}
            \item Gather instructions and data into input string: \gatherinput{$[\text{list of instructions}]$, $[\text{list of data}]$}
            \item LLMs generate API: \gen{messages, temperature, max tokens}
            \item Depth-first search for tertiary and inferior processes: \dfs{dominant process, auxiliary process} 
        \end{enumerate}
        
        \textbf{Step 1:} Dominant process prediction - We ask the model to predict the dominant process based on human features and process descriptions.\\
        \begin{enumerate}[itemsep=0pt, topsep=0pt]
            \item $\text{proc\_desc} \leftarrow$ \getdesc{}\;
            \item $\text{user\_input}_{\text{dom}}\leftarrow$ \gatherinput{$[\text{"", "Stress level"}]$, $[P_{\text{demo}}, S_{\text{overall}_j}]$}\;
            \item $\text{message}\leftarrow$ [\{"role": "system", \\"content": $P_{\text{sys}}$\}, \{"role": "user", "content": $\text{user\_input}_{\text{dom}}$\}, \{"role": "tool", "content": proc\_desc\}]\;
            \item $p_{\text{dom}}\leftarrow$ \gen{\text{message}, $t$, $m_{\text{token}}$}\;
        \end{enumerate}

        \textbf{Step 2:} Auxiliary process prediction - We ask the model to predict the auxiliary process based on human features and candidate process descriptions.\\
        \begin{enumerate}[itemsep=0pt, topsep=0pt]
            \item $\text{aux\_candidates} \leftarrow$ \can{"dominant", $p_{\text{dom}}$}\;
            \item $\text{message}\leftarrow \text{message}+$ [\{"role": "assistant", "content": $p_{\text{dom}}$\}, \{"role": "user", "content": "Select auxiliary from \\process candidates."\}, \{"role": "tool", "content": aux\_candidates\}]\;
            \item $p_{\text{aux}}\leftarrow$ \gen{\text{message}, $t$, $m_{\text{token}}$}\;
        \end{enumerate}
        
        \textbf{Step 3:} Filling the rest processes - We automatically find the tertiary and inferior processes by depth-first search over the process tree (Fig.\ref{fig:funtotype}).\\
        \begin{enumerate}[itemsep=0pt, topsep=0pt]
            \item $p_{\text{ter}}, p_{\text{inf}}, T_j \leftarrow$ \dfs{$p_{\text{dom}}$, $p_{\text{aux}}$}\;
            \item $p_{\text{cog}_j}\leftarrow [p_{\text{dom}}, p_{\text{aux}}, p_{\text{ter}}, p_{\text{inf}}]$\;
        \end{enumerate}
        return $p_{\text{cog}_j}, T_j$
\end{algorithm*}
\begin{table*}[]
\centering
\resizebox{\textwidth}{!}{%
\begin{tabular}{lllllllllllll}
\specialrule{.2em}{.1em}{.1em}
\textbf{System Prompt for Data Augmentation} \\ \hline 
\rowcolor{gray!30} 
\textbf{\textit{Personality Augmentation}} \\
\multicolumn{1}{l}{\begin{tabular}[c]{@{}l@{}}\textbf{Task:} You will analyze the likelihood of different personality types selecting specific choices in response to a survey question. Based on\\ the descriptions of each personality type, predict the probability (as a percentage from 0\% to 100\%) that each type would choose a given option.\end{tabular}} \\
\multicolumn{1}{l}{\textbf{Input Format:}} \\
\multicolumn{1}{l}{Survey Question:} \\
\multicolumn{1}{l}{{~~[}QUESTION{]}} \\
\multicolumn{1}{l}{Choices:} \\
\multicolumn{1}{l}{{~~[}CHOICE 1{]}} \\
\multicolumn{1}{l}{{~~[}CHOICE 2{]}} \\
\multicolumn{1}{l}{{~~[}Additional choices as needed.{]}} \\
\multicolumn{1}{l}{\textbf{Output Format:}} \\
\multicolumn{1}{l}{Provide the predicted probabilities in the following structured format:} \\
\multicolumn{1}{l}{\{} \\
\multicolumn{1}{l}{~~"{[}ROLE/TYPE NAME 1{]}":   "PROB\%",} \\
\multicolumn{1}{l}{~~"{[}ROLE/TYPE NAME 2{]}":   "PROB\%",} \\
\multicolumn{1}{l}{~~...} \\
\multicolumn{1}{l}{\}} \\
\multicolumn{1}{l}{\begin{tabular}[c]{@{}l@{}}Using the provided personality type descriptions, predict the probability (0\% to 100\%) that each type would select each choice in response to the \\      survey question.\end{tabular}} \\ \hline
\rowcolor{gray!30} 
\textbf{\textit{Cultural Value Orientation Augmentation}}\\
\begin{tabular}[c]{@{}l@{}} \textbf{Task:} You are an expert in social science specializing in cross-cultural value studies. Your task is to analyze survey questions, identify the \\ relevant value dimension(s), and determine which answer choices indicate a high score in those dimensions.\end{tabular} \\
\begin{tabular}[c]{@{}l@{}}\textbf{Input:} A survey question with multiple choice answers, pre-assigned value dimension(s), and features of people with high score on \\ this value.\end{tabular} \\
\textbf{Output:} A JSON object containing the question, value dimension(s), and high-scoring choices, formatted as follows: \\
{[}\{"value\_name": "VALUE1", "high\_score\_choices": {[}"CHOICE1", "CHOICE2", ...{]}\}, \\
~~\{"value\_name": "VALUE2", "high\_score\_choices": {[}"CHOICE4", "CHOICE5", ...{]}\}, ...{]} \\
\textbf{Instructions:} \\
1. Carefully read and analyze the given survey question and its answer choices. \\
2. Confirm the relevance of the pre-assigned value dimension(s) to the question. \\
3. Identify which answer choice(s) indicate a high score in the given value dimension(s). \\
4. Ensure each value dimension appears only once in the "values" array. \\
5. Include all relevant high-scoring choices in the "high\_score\_choices" array. \\
6. Ensure your response is in valid JSON format that can be parsed by Python's json package. \\
\begin{tabular}[c]{@{}l@{}}\textbf{Note: }High-scoring choices are those that strongly align with or exemplify the given value dimension(s). Multiple choices may \\ indicate high scores if they all strongly represent the value(s).\end{tabular} \\
\textbf{Example input:} \\
Question: "How important is it for you to follow traditions?" \\
Choices: A) Very important, B) Somewhat important, C) Not very important, D) Not at all important \\
Pre-assigned value: Tradition \\
\begin{tabular}[c]{@{}l@{}}Features of people with high score: At the individual level, those who value tradition highly believe following their culture’s \\ traditions provides meaning, belonging, and guidance in life. They have pride in their heritage and faithfully observe rituals and customs.\end{tabular} \\
\textbf{Example output:} \\
{[}\{"value\_name": "Tradition Index", "high\_score\_choices": {[}"A"{]}\}{]}, \\
Please provide your analysis for the given survey question, choices, and pre-assigned value dimension(s). \\ \hline
\end{tabular}%
}
\caption{Data augmentation prompt for personality and cultural value orientation.}
\label{tbl:aug_prompt}
\end{table*}

% Stage 1 Stress Analysis

\begin{table*}[]
\resizebox{\textwidth}{!}{%
\begin{tabular}{l}
\specialrule{.2em}{.1em}{.1em}
\textbf{System Prompt for Human Stress Analysis Stage} \\ \hline 
\rowcolor{gray!30} 
\textbf{\textit{Stress Level Assign Based on Features}} \\
\begin{tabular}[c]{@{}l@{}}\textbf{Role}: You are a component in a multi-agent system designed to simulate human value preferences based on given features. \\ Your task is to assign stress levels (integers from 0 to 100) to each feature, reflecting its potential impact on the individual's stress. \\These stress levels will be used by other models to simulate the individual's thinking process.\end{tabular} \\
\textbf{Input}: A list of human feature in the format {[}\{feature KEY: VALUE\}{]}.                                                                                                                                                                                                                                                                                                                                           \\
\textbf{Output}: A JSON object containing each features and its assigned stress level, formatted as follows:                                                                                                                                                                                                                                                                                                                      \\
\{"features": {[}                                                                                                                                                                                                                                                                                                                                                                                                            \\
~~\{"features": "feature1", "value": "VALUE1", "stress\_level": INTEGER1, "explanation": "EXPLANATION1"\},                                                                                                                                                                                                                                                                                                               \\
~~\{"features": "feature2", "value": "VALUE2", "stress\_level": INTEGER2, "explanation": "EXPLANATION2"\},                                                                                                                                                                                                                                                                                                               \\
~~...                                                                                                                                                                                                                                                                                                                                                                                                                            \\
{]}\}                                                                                                                                                                                                                                                                                                                                                                                                                          \\
\textbf{Instructions}:                                                                                                                                                                                                                                                                                                                                                                                                                  \\
1. Analyze each features individually.                                                                                                                                                                                                                                                                                                                                                                                   \\
2. Consider the \textit{general impact} of each features on stress levels, not just in specific scenarios.                                                                                                                                                                                                                                                                                                                        \\
3. Assign a stress level (0-100) to each features based on its potential to cause stress.                                                                                                                                                                                                                                                                                                                                \\
4. Ensure each features appears only once in the output, and the output can be directly loaded by JSON package in Python.                                                                                                                                                                                                                                                                                                \\
5. Provide a brief explanation (1-2 sentences) for each assigned stress level.                                                                                                                                                                                                                                                                                                                                                 \\
\begin{tabular}[c]{@{}l@{}}\textbf{Note}: The stress levels you assign will guide a reasoning model in simulating human thinking processes when answering social surveys.\\ Therefore, your analysis should be general and applicable to a wide range of situations.\end{tabular}                                                                                                                                                       \\
\textbf{Example input}:                                                                                                                                                                                                                                                                                                                                                                                                                 \\
{[}\{"income level": "Medium"\}, \{"employment status": "Full time (30 hours a week or more)"\}{]}                                                                                                                                                                                                                                                                                                                             \\
\textbf{Example output}:                                                                                                                                                                                                                                                                                                                                                                                                                \\
\{"features": {[}                                                                                                                                                                                                                                                                                                                                                                                                            \\
\begin{tabular}[c]{@{}l@{}}~~\{"features": "income level", "value": "Medium", "stress\_level": 50, "explanation": "A medium income level can provide stability \\ ~~but may also cause stress due to financial limitations."\},\end{tabular}                                                                                                                                                                                 \\
\begin{tabular}[c]{@{}l@{}}~~\{"features": "employment status", "value": "Full time (30 hours a week or more)", "stress\_level": 60, "explanation": "Full-time \\ ~~employment offers security but can lead to work-related stress and time management issues."\}\end{tabular}                                                                                                                                               \\
{]}\}                                                                                                                                                                                                                                                                                                                                                                                                                          \\
Please proceed with your analysis for the given feature.                                                                                                                                                                                                                                                                                                                                                              \\ \hline
\rowcolor{gray!30} 
\textbf{\textit{Sociodemographic Prompt Generation}} \\
\textbf{Role}: feature Impact Analyzer for Personality Simulation Systems                                    \\
\textbf{Task}: Filter and summarize feature profiles based on their stress impact and personality influence. \\
\textbf{Input Schema}:                                                                                         \\
Human features:                                                                                     \\
\{
"features": {[}
\{
"features": string, 
"value": string, 
"stress\_level": integer(1-100), 
"explanation": string
\}
{]}
\}                                                                                                    \\
Average stress level: xx/100                                                                          \\
\textbf{Evaluation Criteria}:                                                                                  \\
1. Primary Retention Factors:                                                                         \\
~~- Stress level \textit{larger than} the average stress level                                                             \\
~~- Direct impact on simulation outcomes                                                                \\
~~- Critical personality interactions                                                                   \\
2. Secondary Considerations:                                                                          \\
~~- Relationship dependencies                                                                           \\
~~- Value contribution weight                                                                           \\
~~- feature complexity                                                                                \\
\textbf{Output Schema}:                                                                                        \\
\{                                                                                                    \\
~~"kept\_features": {[}\{
"features": string, 
"value": string, 
"stress\_level": integer(1-100), 
"explanation": string, 
"retention\_reason": string
\}{]},                                                                                                  \\
~~"dropped\_features": {[}\{
"features": string, 
"value": string, 
"stress\_level": integer(1-100), 
"exclusion\_reason": string
\}{]},                                                                                                  \\
~~"dropped\_profile": string,                                                                           \\
~~"metadata": \{
"total\_features": integer,
"retained\_count": integer,
"average\_stress\_retained": float,
\} \\
\}                                                                                                    \\
\textbf{Requirements}:                                                                                         \\
- Maintain original data structure for retained features                                            \\
- Provide clear exclusion reasoning                                                                   \\
- Write a one-paragraph profile describing the human features based on dropped feature values       \\
- Ensure traceability between related features                                                      \\
- Only output the JSON results and make sure the key are the same as presented in output schema       \\
Please proceed with your summarization of the given feature.      \\ \hline                           
\end{tabular}%
}
\caption{System prompt for human stress analysis.}
\label{tbl:fea_ana_prompt}
\end{table*}

% Stage 2 Personality Prediction

\begin{table*}[]
\resizebox{\textwidth}{!}{%
\begin{tabular}{l}
\specialrule{.2em}{.1em}{.1em}
\textbf{System Prompt for Personality Prediction Stage} \\ \hline 
\rowcolor{gray!30} 
\textbf{\textit{Single Process Prediction}} \\
\begin{tabular}[c]{@{}l@{}}\textbf{Role}: You are the initial node in a multi-agent system designed to simulate human value preferences based on provided human feature. \\ Your task is to assess the current stage of reasoning and select the appropriate thinking process from the 8 processes defined by the MBTI \\ Processes of Type Dynamics theory.\end{tabular} \\
\textbf{Input}:                                                                                                                                                                                                                                                                                                                                                                   \\
1. Human feature                                                                                                                                                                                                                                                                                                                                                 \\
2. Current stress level                                                                                                                                                                                                                                                                                                                                                  \\
3. Current reasoning step and selected thinking process                                                                                                                                                                                                                                                                                                                  \\
\textbf{Output}: A JSON array containing the selected reasoning processes and current reasoning stages.                                                                                                                                                                                                                                                                           \\
\textbf{Instructions}:                                                                                                                                                                                                                                                                                                                                                            \\
1. Conduct reasoning in four stages according to the MBTI Processes of Type Dynamics theory:                                                                                                                                                                                                                                                                             \\
\begin{tabular}[c]{@{}l@{}}~~a. Dominant process: The Dominant Process in MBTI type dynamics is the most developed, conscious, and comfortable aspect of an individual's \\ ~~personality, serving as their core and primary cognitive function.\end{tabular}                                                                                                                \\
\begin{tabular}[c]{@{}l@{}}~~b. Auxiliary process: The Auxiliary Process in MBTI type dynamics is the second most developed function that supports and balances the \\ ~~Dominant Process, aiding in perception, judgment, and the balance between Extraversion and Introversion.\end{tabular}                                                                               \\
\begin{tabular}[c]{@{}l@{}}~~c. Tertiary process: The Tertiary Process in MBTI type dynamics is the third most developed function, less competent and comfortable than the \\ ~~Dominant and Auxiliary, typically activated under stress for decision-making and problem-solving.\end{tabular}                                                                               \\
\begin{tabular}[c]{@{}l@{}}~~d. Inferior process: The Inferior Process in MBTI type dynamics is the least developed function, emerging under extreme stress and potentially \\ ~~causing individuals to feel out of control or display negative feature.\end{tabular}                                                                                                \\
2. For each stage:                                                                                                                                                                                                                                                                                                                                                       \\
~~a. Use the "\textit{get\_all\_process\_desc}" tool to access all the thinking processes' descriptions.                                                                                                                                                                                                                                                                            \\
~~b. Select the appropriate process based on the human feature.                                                                                                                                                                                                                                                                                                    \\
~~c. Use the "\textit{get\_next\_process}" tool to access process candidates for the next stage.                                                                                                                                                                                                                                                                                    \\
4. Use the provided tools for additional information as needed.                                                                                                                                                                                                                                                                                                          \\
5. Only output the JSON results and make sure the key are the same as presented in example output.                                                                                                                                                                                                                                                                       \\
\textbf{Tools}:                                                                                                                                                                                                                                                                                                                                                                   \\
{[}\{"tool\_name": "get\_all\_process\_desc", "description": "obtain all the thinking process descriptions", "inputs": None\},                                                                                                                                                                                                                                           \\
\begin{tabular}[c]{@{}l@{}}~~\{"tool\_name": "get\_next\_process", "description": "access process candidates for the next stage based on current process", "input": \{"current\_stage": \\ ~~CURRENT\_STAGE, "previous\_processes":{[}PROCESS1, PROCESS2, 
\\~~...
{]}\} \\
\}{]}\end{tabular}                                                                                         \\
\textbf{Note}: Ensure each process appears only once in the output.                                                                                                                                                                                                                                                                                                               \\
\textbf{Example output}:                                                                                                                                                                                                                                                                                                                                                          \\
{[}\{
"reasoning\_stage": "Dominant",
"process": "Extraverted Thinking",
\},                                                                                                                                                                                                                                                                                                                                                                      \\
~~\{
"reasoning\_stage": "Auxiliary",
"process": "Introverted Intuition",
\},\\
~~... \\
{]}                                                                                                                                                                                                                                                                                                                                                                      \\
Please proceed with your analysis for the given human feature, stress level, and survey questions.                                                                                                                                                                                                                                                             \\ \hline 
\end{tabular}%
}
\caption{System prompt for personality prediction.}
\label{tbl:person_pred_prompt}
\end{table*}

% Stage 3 Cognitive Reasoning

\begin{table*}[]
\resizebox{\textwidth}{!}{%
\begin{tabular}{l}
\specialrule{.2em}{.1em}{.1em}
\textbf{System Prompt for Cognitive Reasoning Stage} \\ \hline 
\rowcolor{gray!30} 
\textbf{\textit{Assign Stress Impact on Processes}} \\
\begin{tabular}[c]{@{}l@{}}\textbf{Role}: You are the second node in a multi-agent system designed to simulate human value preferences based on provided human feature.\\ Your task is to assess each of the reasoning stages and assign the appropriate stress impact based on survey questions and feature-based stress level.\end{tabular} \\
\textbf{Input}:                                                                                                                                                                                                                                                                                                                                       \\
1. Human feature                                                                                                                                                                                                                                                                                                                     \\
2. Current stress level                                                                                                                                                                                                                                                                                                                      \\
3. Selected thinking process                                                                                                                                                                                                                                                                                                                 \\
4. Survey questions                                                                                                                                                                                                                                                                                                                          \\
\textbf{Output}: A JSON array containing the selected reasoning processes and current reasoning stages.                                                                                                                                                                                                                                               \\
\textbf{Instructions}:                                                                                                                                                                                                                                                                                                                                \\
1. For each stage, during evaluating the stress impact:                                                                                                                                                                                                                                                                                      \\
~~- Assess if the survey questions may increase or decrease stress.                                                                                                                                                                                                                                                                            \\
~~- Determine if the stress level is high enough to \textit{trigger the negative aspect} of the process.                                                                                                                                                                                                                                                \\
~~- Output "negative" if triggered, otherwise "positive".                                                                                                                                                                                                                                                                                      \\
2. Only output the JSON results and make sure the key is the same as presented in the example output.                                                                                                                                                                                                                                           \\
\textbf{Note}: Ensure each process appears only once in the output.                                                                                                                                                                                                                                                                                   \\
\textbf{Example output}:                                                                                                                                                                                                                                                                                                                              \\
{[}                                                                                                                                                                                                                                                                                                                                          \\
~~\{
"reasoning\_stage": "Dominant",
"process": "Extraverted Thinking",
"stress\_impact": "positive"
\},                                                                                                                                                                                                                                                                                                                                          \\
~~\{
"reasoning\_stage": "Auxiliary",
"process": "Introverted Intuition",
"stress\_impact": "negative"
\},                                                                                                                                                                                                                                                                                                                                          \\
~~...                                                                                                                                                                                                                                                                                                                                          \\
{]}                                                                                                                                                                                                                                                                                                                                          \\
Please proceed with your analysis of the given human feature, stress level, and survey questions.                                                                                                                                                                                                                                  \\ \hline
\rowcolor{gray!30} 
\textbf{\textit{Predict Answer with Four Processes}} \\
\begin{tabular}[c]{@{}l@{}}\textbf{Role}: You are the second node in a multi-agent system designed to simulate human value preferences based on provided human feature.\\  Your task is to complete the reasoning processes given by the previous node and answer survey questions based on the human feature and \\ stress impacts on each process.\end{tabular} \\
\textbf{Input}:                                                                                                                          \\
1. Human feature                                                                                                        \\
2. Overall stress level                                                                                                         \\
3. Survey questions with multiple-choice options                                                                                 \\
4. Previous node's output (reasoning processes and stress impacts)                                                              \\
\textbf{Output}: A JSON array containing the reasoning results for each process.                                                         \\
\textbf{Instructions}:                                                                                                                   \\
1. For each of the four MBTI Processes of Type Dynamics theory stages:                                                          \\
~~a. Conduct reasoning according to the process description, human feature and answer the survey question.                \\
~~b. Provide the reasoning result for each process and a brief explanation (1-2 sentences).                                       \\
~~c. Assign a weight (0 to 1) to each process based on the overall stress level and the process's position in the MBTI hierarchy. \\
2. Only output the JSON results and make sure the key are the same as presented in example output.                              \\
\textbf{Note}: Ensure each process appears only once in the output.                                                                      \\
\textbf{Example output}:                                                                                                                 \\
{[}\{ 
\\
~~"reasoning\_stage": "Dominant",
"process": "Extraverted Thinking",                                                            \\
~~"process\_description": "Organizing and sequencing information and ideas",                                                      \\
~~"reasoning\_result": "(B) somewhat agree",                                                                                      \\
~~"reasoning\_explained": "The individual is likely to approach this question by logically analyzing the options and their potential consequences.", \\
~~"weight": 0.4                                                                                                                   \\
\}, ... {]}                                                                                                                             \\
Please proceed with your analysis for the given human feature, stress levels, and survey question. \\
\hline
\end{tabular}%
}
\caption{System prompt for cognitive reasoning.}
\label{tbl:cog_prompt}
\end{table*}

% Stage 4 Synthesis

\begin{table*}[]
\resizebox{\textwidth}{!}{%
\begin{tabular}{l}
\specialrule{.2em}{.1em}{.1em}
\textbf{System Prompt for Synthesis Stage} \\ \hline 
\rowcolor{gray!30} 
\textbf{\textit{Synthesize Process Results}} \\
\begin{tabular}[c]{@{}l@{}}\textbf{Role}: You are the final node in a multi-agent system designed to simulate human value preferences based on provided human feature \\ and thinking processes. Your task is to evaluate the reasoning results from previous nodes, ensure their alignment with process descriptions, \\ and provide a final decision based on weighted results.\end{tabular} \\
\textbf{Input}:                                                                                                                                               \\
1. Human feature                                                                                                                             \\
2. Survey question with multiple-choice options                                                                                                      \\
3. Previous nodes' output (reasoning processes, descriptions, results, and weights)                                                                  \\
\textbf{Output}: A JSON array containing the evaluated reasoning results for each process and a final conclusion.                                             \\
\textbf{Instructions}:                                                                                                                                        \\
1. For each of the four MBTI Processes of Type Dynamics theory stages:                                                                               \\
~~a. Review the reasoning result and explanation provided by the previous node.                                                                        \\
~~b. Verify if the reasoning aligns with the process description and given explanation.                                                                \\
~~c. If misaligned, rewrite the result and explanation to ensure consistency.                                                                          \\
~~d. Ensure the weight assigned is appropriate for the process's position in the MBTI hierarchy and the overall stress level.                          \\
2. After evaluating all four processes:                                                                                                              \\
~~a. Synthesize the reasoning results, considering the weights assigned to each process.                                                               \\
~~b. Choose the most appropriate answer from the given survey options based on the weighted synthesis.                                                 \\
\begin{tabular}[c]{@{}l@{}}~~c. Provide a brief explanation (1-2 sentences) for the final decision, explicitly mentioning how the weights of each process influenced the \\ ~~conclusion.\end{tabular} \\
3. Format the output as a JSON array, including all process evaluations and the final conclusion.                                                    \\
4. Only output the JSON results and make sure the key are the same as presented in example output                                                    \\
\textbf{Note}: Ensure each process appears only once in the output, and the conclusion matches one of the given survey question choices.                      \\
\textbf{Example output}:                                                                                                                                      \\
{[} \\
~~\{ "reasoning\_stage": "Dominant",
"process": "Extraverted Thinking",                                                                                                                   \\
~~"process\_description": "Organizing and sequencing information and ideas",                                                                           \\
~~"reasoning\_result": "(B) somewhat agree",                                                                                                           \\
~~"reasoning\_explanation": "The individual is likely to approach this question by logically analyzing the options and their potential consequences.", \\
~~"reasoning\_evaluate": "This aligns with the Extraverted Thinking process, which focuses on external logic and organization of ideas.",              \\
~~"weight": 0.4                                                                                                                                        \\
~~\}, ...,                                                                                                                                                 \\
~~\{ "conclusion": "(B) Somewhat agree",                                                                                                                  \\
\begin{tabular}[c]{@{}l@{}}~~"explanation": "The final decision leans towards 'Somewhat agree' due to the dominant Extraverted Thinking process (40\% weight) favoring a \\ ~~logical approach, balanced by the auxiliary Introverted Intuition (30\% weight) considering long-term implications. The tertiary and inferior \\ ~~processes (15\% each) provided minor influences, resulting in a measured agreement rather than a strong stance."\end{tabular} \\
~~\}                                                                                                                                                   \\
{]}                                                                                                                                                  \\
Please proceed with your analysis and final decision based on the given reasoning results and weights. \\ \hline
\end{tabular}%
}
\caption{System prompt for synthesis stage.}
\label{tbl:syn_prompt}
\end{table*}

\begin{table*}[]
\centering
\resizebox{\textwidth}{!}{%
\begin{tabular}{c|cccc|cccc}
\toprule
\textbf{Model} & \multicolumn{4}{c|}{\textbf{GLM-4-air}} & \multicolumn{4}{c}{\textbf{Doubao-1.5-pro}} \\ \midrule
\textbf{Method} & \textbf{Demo.+Ideo.} & \textbf{\begin{tabular}[c]{@{}c@{}}Demo.+Ideo.+\\ Opiniontop3\end{tabular}} & \textbf{\citet{zhao-etal-2024-worldvaluesbench}} & \textbf{MARK (Ours)} & \textbf{Demo.+Ideo.} & \textbf{\begin{tabular}[c]{@{}c@{}}Demo.+Ideo.+\\ Opiniontop3\end{tabular}} & \textbf{\citet{zhao-etal-2024-worldvaluesbench}} & \textbf{MARK (Ours)} \\ \midrule
$C_0$ & 24.24 & 23.64 & 22.72 & \textbf{34.28\%} & 23.48 & 25.12 & 23.48 & \textbf{42.20} \\
$C_1$ & 19.47 & 26.25 & 28.33 & \textbf{34.18\%} & 26.61 & 27.53 & 22.00 & \textbf{47.82} \\
$C_2$ & 16.01 & 27.24 & 31.67 & \textbf{34.01\%} & 29.63 & 29.67 & 22.59 & \textbf{49.82} \\
$C_3$ & 28.54 & 27.76 & 29.81 & \textbf{33.99\%} & 37.47 & 34.98 & 28.99 & \textbf{51.11} \\
$C_4$ & 32.24 & 29.77 & 34.09 & 33.91\% & 38.09 & 37.41 & 25.18 & \textbf{49.84} \\
$C_5$ & 27.49 & 31.34 & 31.40 & \textbf{34.59\%} & 30.70 & 31.93 & 22.10 & \textbf{45.21} \\
$C_6$ & 28.44 & 27.41 & 27.37 & \textbf{33.62\%} & 25.95 & 26.83 & 25.62 & \textbf{44.20} \\
$C_7$ & 23.21 & 28.17 & 22.65 & \textbf{33.35\%} & 35.06 & 36.60 & 20.45 & \textbf{44.64} \\
$C_8$ & 23.33 & 32.61 & 21.79 & \textbf{34.01\%} & 24.82 & 25.53 & 24.46 & \textbf{41.13} \\
$C_9$ & 25.51 & 21.98 & 27.16 & \textbf{33.44\%} & 25.97 & 27.72 & 23.13 & \textbf{52.06} \\
$C_{10}$ & 29.34 & 32.55 & 33.37 & \textbf{33.42\%} & 41.42 & 42.14 & 28.68 & \textbf{56.96} \\
$C_{11}$ & 28.93 & 27.49 & 30.88 & \textbf{33.87\%} & 38.64 & 37.47 & 24.51 & \textbf{46.69} \\
$C_{12}$ & 8.83 & 6.13 & 8.77 & \textbf{33.05\%} & 31.32 & 30.80 & 20.00 & \textbf{41.05} \\
$C_{13}$ & 29.20 & 18.70 & 26.07 & \textbf{33.60\%} & 24.59 & 26.13 & 32.45 & \textbf{41.85} \\
$C_{14}$ & 30.70 & 27.26 & 28.02 & \textbf{33.09\%} & 28.50 & 30.35 & 23.04 & \textbf{47.74} \\
$C_{15}$ & 26.73 & 22.80 & 29.46 & \textbf{33.42\%} & 36.60 & 36.23 & 29.51 & \textbf{49.44} \\
$C_{16}$ & 25.00 & 19.90 & 22.94 & \textbf{33.75\%} & 29.86 & 32.98 & 19.96 & \textbf{50.64} \\
$C_{17}$ & 29.32 & 21.61 & 26.98 & \textbf{33.62\%} & 20.97 & 24.46 & 18.64 & \textbf{39.53} \\
$C_{18}$ & 28.19 & 23.66 & 29.86 & \textbf{33.58\%} & 33.56 & 31.98 & 26.26 & \textbf{45.14} \\
$C_{19}$ & 25.10 & 24.84 & 26.25 & \textbf{33.02\%} & 31.65 & 33.19 & 23.56 & \textbf{52.45} \\ \midrule
Avg. & 25.49 & 25.06 & 26.98 & \textbf{33.69\%} & 30.75 & 31.45 & 24.23 & \textbf{46.98} \\ \bottomrule
\end{tabular}%
}
\caption{The simulation ACC for U.S. culture. Our method achieves the highest ACC (\%) on some clusters for both models.}
\label{tbl:full_result}
\end{table*}

\begin{table*}[]
\resizebox{\textwidth}{!}{%
\begin{tabular}{c|cccc|cccc}
\toprule
\textbf{Model} & \multicolumn{4}{c|}{\textbf{GLM-4-air}} & \multicolumn{4}{c}{\textbf{Doubao-1.5-pro}} \\ \midrule
\textbf{Method} & \textbf{Demo.+Ideo.} & \textbf{\begin{tabular}[c]{@{}c@{}}Demo.+Ideo.+\\ Opiniontop3\end{tabular}} & \textbf{\citet{zhao-etal-2024-worldvaluesbench}} & \textbf{MARK (Ours)} & \textbf{Demo.+Ideo.} & \textbf{\begin{tabular}[c]{@{}c@{}}Demo.+Ideo.+\\ Opiniontop3\end{tabular}} & \textbf{\citet{zhao-etal-2024-worldvaluesbench}} & \textbf{MARK (Ours)} \\ \midrule
$C_0$ & 0.3630 & 0.4189 & 0.3523 & \textbf{0.4248} & 0.4167 & 0.4481 & 0.3586 & \textbf{0.5083} \\
$C_1$ & 0.3364 & 0.4517 & 0.3933 & 0.4445 & 0.4231 & 0.4551 & 0.3452 & \textbf{0.5462} \\
$C_2$ & 0.3182 & 0.4251 & 0.4135 & \textbf{0.4503} & 0.4343 & 0.4585 & 0.3400 & \textbf{0.5447} \\
$C_3$ & 0.3423 & 0.4438 & 0.4038 & \textbf{0.4665} & 0.4818 & 0.4945 & 0.3916 & \textbf{0.5610} \\
$C_4$ & 0.3667 & 0.4662 & 0.4451 & \textbf{0.4718} & 0.4654 & 0.4735 & 0.3759 & \textbf{0.5409} \\
$C_5$ & 0.3582 & 0.4617 & 0.4239 & 0.4447 & 0.4709 & 0.4805 & 0.3528 & \textbf{0.5040} \\
$C_6$ & 0.3620 & 0.4580 & 0.3913 & 0.4518 & 0.4768 & 0.4897 & 0.3500 & \textbf{0.5388} \\
$C_7$ & 0.3571 & 0.4332 & 0.3523 & 0.4219 & 0.4650 & 0.4803 & 0.3381 & \textbf{0.5055} \\
$C_8$ & 0.3353 & 0.4584 & 0.3480 & \textbf{0.4887} & 0.4280 & 0.4439 & 0.3545 & \textbf{0.4767} \\
$C_9$ & 0.3614 & 0.4317 & 0.3818 & 0.4374 & 0.4726 & 0.4844 & 0.3637 & \textbf{0.5388} \\
$C_{10}$ & 0.3794 & 0.4317 & 0.4120 & \textbf{0.4609} & 0.5030 & 0.5254 & 0.3841 & \textbf{0.5683} \\
$C_{11}$ & 0.3614 & 0.4496 & 0.3907 & 0.4450 & 0.4625 & 0.4692 & 0.3469 & \textbf{0.5055} \\
$C_{12}$ & 0.2316 & 0.2623 & 0.2368 & \textbf{0.2869} & 0.4204 & 0.4367 & 0.3312 & \textbf{0.4654} \\
$C_{13}$ & 0.3646 & 0.4177 & 0.3706 & 0.4063 & 0.4304 & 0.4418 & 0.3863 & \textbf{0.4775} \\
$C_{14}$ & 0.3753 & 0.4482 & 0.4015 & 0.4412 & 0.4811 & 0.4927 & 0.3682 & \textbf{0.5346} \\
$C_{15}$ & 0.3761 & 0.4302 & 0.3895 & \textbf{0.4473} & 0.4807 & 0.4904 & 0.3743 & \textbf{0.5339} \\
$C_{16}$ & 0.3458 & 0.4194 & 0.3581 & \textbf{0.4339} & 0.4571 & 0.4757 & 0.3459 & \textbf{0.5361} \\
$C_{17}$ & 0.3839 & 0.4333 & 0.3897 & 0.4020 & 0.4133 & 0.4339 & 0.3336 & \textbf{0.4563} \\
$C_{18}$ & 0.3875 & 0.4376 & 0.3904 & 0.4282 & 0.4685 & 0.4752 & 0.3630 & \textbf{0.5096} \\
$C_{19}$ & 0.3722 & 0.4468 & 0.3824 & 0.4411 & 0.4751 & 0.4975 & 0.3630 & \textbf{0.5388} \\ \midrule
Avg. & 0.3539 & 0.4313 & 0.3814 & \textbf{0.4348} & 0.4563 & 0.4723 & 0.3584 & \textbf{0.5195} \\ \bottomrule
\end{tabular}%
}
\caption{The simulation distribution divergence for U.S. culture. Our method achieves a higher 1-JSD score on some clusters for feature clustering.}
\label{tbl:full_jsd}
\end{table*}

\begin{table*}[]
\resizebox{\textwidth}{!}{%
\begin{tabular}{c|cccc|cccc}
\toprule
\textbf{Model} & \multicolumn{4}{c|}{\textbf{GLM-4-air}} & \multicolumn{4}{c}{\textbf{Doubao-1.5-pro}} \\ \midrule
\textbf{Method} & \textbf{Demo.+Ideo.} & \textbf{\begin{tabular}[c]{@{}c@{}}Demo.+Ideo.+\\ Opiniontop3\end{tabular}} & \textbf{\citet{zhao-etal-2024-worldvaluesbench}} & \textbf{MARK (Ours)} & \textbf{Demo.+Ideo.} & \textbf{\begin{tabular}[c]{@{}c@{}}Demo.+Ideo.+\\ Opiniontop3\end{tabular}} & \textbf{\citet{zhao-etal-2024-worldvaluesbench}} & \textbf{MARK (Ours)} \\ \midrule
$C_0$ & 0.0330 & 0.0578 & 0.0392 & 0.0821 & 0.0293 & 0.0360 & 0.0182 & 0.0338 \\
$C_1$ & 0.0455 & 0.0682 & 0.0419 & 0.0874 & 0.0597 & 0.0596 & 0.0246 & 0.0532 \\
$C_2$ & 0.0694 & 0.0697 & 0.0424 & 0.0870 & 0.0660 & 0.0640 & 0.0295 & 0.0527 \\
$C_3$ & 0.0745 & 0.0648 & 0.0427 & 0.0861 & 0.0706 & 0.0694 & 0.0326 & 0.0560 \\
$C_4$ & 0.0780 & 0.0640 & 0.0426 & 0.0858 & 0.0678 & 0.0649 & 0.0305 & 0.0547 \\
$C_5$ & 0.0792 & 0.0612 & 0.0439 & 0.0859 & 0.0669 & 0.0650 & 0.0291 & 0.0543 \\
$C_6$ & 0.0785 & 0.0648 & 0.0485 & 0.0868 & 0.0681 & 0.0680 & 0.0382 & 0.0584 \\
$C_7$ & 0.0786 & 0.0652 & 0.0476 & 0.0879 & 0.0671 & 0.0688 & 0.0381 & 0.0591 \\
$C_8$ & 0.0768 & 0.0654 & 0.0470 & 0.0885 & 0.0707 & 0.0732 & 0.0387 & 0.0610 \\
$C_9$ & 0.0788 & 0.0685 & 0.0461 & 0.0893 & 0.0729 & 0.0749 & 0.0369 & 0.0600 \\
$C_{10}$ & 0.0792 & 0.0685 & 0.0458 & 0.0900 & 0.0703 & 0.0737 & 0.0378 & 0.0599 \\
$C_{11}$ & 0.0786 & 0.0671 & 0.0457 & 0.0902 & 0.0729 & 0.0771 & 0.0389 & 0.0587 \\
$C_{12}$ & 0.0796 & 0.0660 & 0.0453 & 0.0907 & 0.0705 & 0.0752 & 0.0390 & 0.0574 \\
$C_{13}$ & 0.0802 & 0.0684 & 0.0468 & 0.0907 & 0.0729 & 0.0766 & 0.0428 & 0.0573 \\
$C_{14}$ & 0.0799 & 0.0681 & 0.0470 & 0.0908 & 0.0724 & 0.0756 & 0.0416 & 0.0576 \\
$C_{15}$ & 0.0792 & 0.0686 & 0.0467 & 0.0909 & 0.0718 & 0.0748 & 0.0439 & 0.0574 \\
$C_{16}$ & 0.0783 & 0.0694 & 0.0464 & 0.0910 & 0.0716 & 0.0746 & 0.0426 & 0.0569 \\
$C_{17}$ & 0.0787 & 0.0709 & 0.0465 & 0.0910 & 0.0757 & 0.0781 & 0.0413 & 0.0574 \\
$C_{18}$ & 0.0778 & 0.0704 & 0.0462 & 0.0912 & 0.0752 & 0.0774 & 0.0423 & 0.0572 \\
$C_{19}$ & 0.0783 & 0.0710 & 0.0461 & 0.0916 & 0.0770 & 0.0781 & 0.0423 & 0.0581 \\ \midrule
Avg. & 0.0741 & 0.0669 & 0.0452 & 0.0887 & 0.0685 & 0.0703 & 0.0364 & 0.0561 \\ \bottomrule
\end{tabular}%
}
\caption{The simulation distribution divergence for U.S. culture. MARK achieves a lower EMD score on some clusters for both models against most methods.}
\label{tbl:full_emd}
\end{table*}

\begin{table*}[]
\resizebox{\textwidth}{!}{%
\begin{tabular}{c|cccc|cccc|cccc}
\toprule
\textbf{Metrics} & \multicolumn{4}{c|}{\textbf{ACC (\%)}} & \multicolumn{4}{c|}{\textbf{1-JSD ↑}} & \multicolumn{4}{c}{\textbf{EMD ↓}} \\ \midrule
\textbf{Model} & \multicolumn{2}{c|}{\textbf{GLM-4-air}} & \multicolumn{2}{c|}{\textbf{Doubao-1.5-pro}} & \multicolumn{2}{c|}{\textbf{GLM-4-air}} & \multicolumn{2}{c|}{\textbf{Doubao-1.5-pro}} & \multicolumn{2}{c|}{\textbf{GLM-4-air}} & \multicolumn{2}{c}{\textbf{Doubao-1.5-pro}} \\
\textbf{Method} & \textbf{\begin{tabular}[c]{@{}c@{}}Predicted\\ personality\end{tabular}} & \multicolumn{1}{c|}{\textbf{\begin{tabular}[c]{@{}c@{}}Given\\ personality\end{tabular}}} & \textbf{\begin{tabular}[c]{@{}c@{}}Predicted\\ personality\end{tabular}} & \textbf{\begin{tabular}[c]{@{}c@{}}Given\\ personality\end{tabular}} & \textbf{\begin{tabular}[c]{@{}c@{}}Predicted\\ personality\end{tabular}} & \multicolumn{1}{c|}{\textbf{\begin{tabular}[c]{@{}c@{}}Given\\ personality\end{tabular}}} & \textbf{\begin{tabular}[c]{@{}c@{}}Predicted\\ personality\end{tabular}} & \textbf{\begin{tabular}[c]{@{}c@{}}Given\\ personality\end{tabular}} & \textbf{\begin{tabular}[c]{@{}c@{}}Predicted\\ personality\end{tabular}} & \multicolumn{1}{c|}{\textbf{\begin{tabular}[c]{@{}c@{}}Given\\ personality\end{tabular}}} & \textbf{\begin{tabular}[c]{@{}c@{}}Predicted\\ personality\end{tabular}} & \textbf{\begin{tabular}[c]{@{}c@{}}Given\\ personality\end{tabular}} \\ \midrule
\textbf{$C_0$} & 34.28 & \multicolumn{1}{c|}{32.24} & 42.20 & \textbf{42.84} & 0.4248 & \multicolumn{1}{c|}{0.4281} & 0.5083 & \textbf{0.5131} & 0.0821 & \multicolumn{1}{c|}{\textbf{0.0801}} & 0.0338 & 0.0388 \\
\textbf{$C_1$} & 34.18 & \multicolumn{1}{c|}{\textbf{35.19}} & 47.82 & 47.59 & 0.4445 & \multicolumn{1}{c|}{\textbf{0.4490}} & 0.5462 & \textbf{0.5465} & 0.0874 & \multicolumn{1}{c|}{\textbf{0.0826}} & 0.0532 & \textbf{0.0527} \\
\textbf{$C_2$} & 34.01 & \multicolumn{1}{c|}{\textbf{36.32}} & 49.82 & \textbf{50.21} & 0.4503 & \multicolumn{1}{c|}{\textbf{0.4543}} & 0.5447 & \textbf{0.5508} & 0.0870 & \multicolumn{1}{c|}{\textbf{0.0817}} & 0.0527 & \textbf{0.0511} \\
\textbf{$C_3$} & 33.99 & \multicolumn{1}{c|}{\textbf{38.52}} & 51.11 & \textbf{52.45} & 0.4665 & \multicolumn{1}{c|}{\textbf{0.4711}} & 0.5610 & \textbf{0.5664} & 0.0861 & \multicolumn{1}{c|}{\textbf{0.0823}} & 0.0560 & \textbf{0.0502} \\
\textbf{$C_4$} & 33.91 & \multicolumn{1}{c|}{\textbf{38.48}} & 49.84 & \textbf{49.96} & 0.4718 & \multicolumn{1}{c|}{0.4688} & 0.5409 & \textbf{0.5480} & 0.0858 & \multicolumn{1}{c|}{\textbf{0.0826}} & 0.0547 & \textbf{0.0493} \\
\textbf{$C_5$} & 34.59 & \multicolumn{1}{c|}{\textbf{34.94}} & 45.21 & \textbf{45.45} & 0.4447 & \multicolumn{1}{c|}{\textbf{0.4465}} & 0.5040 & \textbf{0.5087} & 0.0859 & \multicolumn{1}{c|}{\textbf{0.0834}} & 0.0543 & \textbf{0.0497} \\
\textbf{$C_6$} & 33.62 & \multicolumn{1}{c|}{\textbf{35.70}} & 44.20 & \textbf{44.40} & 0.4518 & \multicolumn{1}{c|}{0.4501} & 0.5388 & 0.5396 & 0.0868 & \multicolumn{1}{c|}{\textbf{0.0840}} & 0.0584 & \textbf{0.0545} \\
\textbf{$C_7$} & 33.35 & \multicolumn{1}{c|}{30.58} & 44.64 & \textbf{45.82} & 0.4219 & \multicolumn{1}{c|}{\textbf{0.4228}} & 0.5055 & \textbf{0.5106} & 0.0879 & \multicolumn{1}{c|}{\textbf{0.0861}} & 0.0591 & \textbf{0.0547} \\
\textbf{$C_8$} & 34.01 & \multicolumn{1}{c|}{\textbf{39.77}} & 41.13 & \textbf{41.17} & 0.4887 & \multicolumn{1}{c|}{0.4822} & 0.4767 & 0.4732 & 0.0885 & \multicolumn{1}{c|}{\textbf{0.0863}} & 0.0610 & \textbf{0.0573} \\
\textbf{$C_9$} & 33.44 & \multicolumn{1}{c|}{32.96} & 52.06 & 50.08 & 0.4374 & \multicolumn{1}{c|}{0.4321} & 0.5388 & \textbf{0.5424} & 0.0893 & \multicolumn{1}{c|}{\textbf{0.0858}} & 0.0600 & \textbf{0.0575} \\
\textbf{$C_{10}$} & 33.42 & \multicolumn{1}{c|}{\textbf{35.97}} & 56.96 & 56.63 & 0.4609 & \multicolumn{1}{c|}{0.4602} & 0.5683 & \textbf{0.5738} & 0.0900 & \multicolumn{1}{c|}{\textbf{0.0871}} & 0.0599 & \textbf{0.0572} \\
\textbf{$C_{11}$} & 33.87 & \multicolumn{1}{c|}{\textbf{35.08}} & 46.69 & 45.86 & 0.4450 & \multicolumn{1}{c|}{0.4427} & 0.5055 & \textbf{0.5062} & 0.0902 & \multicolumn{1}{c|}{\textbf{0.0872}} & 0.0587 & \textbf{0.0565} \\
\textbf{$C_{12}$} & 33.05 & \multicolumn{1}{c|}{14.26} & 41.05 & \textbf{41.23} & 0.2869 & \multicolumn{1}{c|}{0.2828} & 0.4654 & \textbf{0.4659} & 0.0907 & \multicolumn{1}{c|}{\textbf{0.0875}} & 0.0574 & \textbf{0.0554} \\
\textbf{$C_{13}$} & 33.60 & \multicolumn{1}{c|}{30.43} & 41.85 & 41.73 & 0.4063 & \multicolumn{1}{c|}{0.4050} & 0.4775 & 0.4777 & 0.0907 & \multicolumn{1}{c|}{\textbf{0.0877}} & 0.0573 & \textbf{0.0559} \\
\textbf{$C_{14}$} & 33.09 & \multicolumn{1}{c|}{\textbf{33.85}} & 47.74 & \textbf{47.98} & 0.4412 & \multicolumn{1}{c|}{0.4387} & 0.5346 & 0.5334 & 0.0908 & \multicolumn{1}{c|}{\textbf{0.0879}} & 0.0576 & \textbf{0.0566} \\
\textbf{$C_{15}$} & 33.42 & \multicolumn{1}{c|}{\textbf{34.36}} & 49.44 & \textbf{49.53} & 0.4473 & \multicolumn{1}{c|}{0.4458} & 0.5339 & \textbf{0.5370} & 0.0909 & \multicolumn{1}{c|}{\textbf{0.0881}} & 0.0574 & \textbf{0.0564} \\
\textbf{$C_{16}$} & 33.75 & \multicolumn{1}{c|}{\textbf{35.37}} & 50.64 & 49.98 & 0.4339 & \multicolumn{1}{c|}{0.4317} & 0.5361 & \textbf{0.5358} & 0.0910 & \multicolumn{1}{c|}{\textbf{0.0883}} & 0.0569 & \textbf{0.0560} \\
\textbf{$C_{17}$} & 33.62 & \multicolumn{1}{c|}{29.55} & 39.53 & 38.27 & 0.4020 & \multicolumn{1}{c|}{\textbf{0.4038}} & 0.4563 & \textbf{0.4570} & 0.0910 & \multicolumn{1}{c|}{\textbf{0.0880}} & 0.0574 & \textbf{0.0564} \\
\textbf{$C_{18}$} & 33.58 & \multicolumn{1}{c|}{\textbf{35.68}} & 45.14 & 44.71 & 0.4282 & \multicolumn{1}{c|}{0.4264} & 0.5096 & 0.5082 & 0.0912 & \multicolumn{1}{c|}{\textbf{0.0881}} & 0.0572 & \textbf{0.0562} \\
\textbf{$C_{19}$} & 33.02 & \multicolumn{1}{c|}{\textbf{33.93}} & 52.45 & 52.06 & 0.4411 & \multicolumn{1}{c|}{0.4406} & 0.5388 & \textbf{0.5393} & 0.0916 & \multicolumn{1}{c|}{\textbf{0.0881}} & 0.0581 & \textbf{0.0572} \\ \midrule
Avg. & 33.69 & \multicolumn{1}{c|}{33.66} & 46.98 & \textbf{46.90} & 0.4348 & \multicolumn{1}{c|}{0.4341} & 0.5195 & \textbf{0.5217} & 0.0887 & \multicolumn{1}{c|}{\textbf{0.0857}} & 0.0561 & \textbf{0.0540} \\ \bottomrule
\end{tabular}%
}
\caption{The simulation performance with predicted personality and accurate personality on the U.S. culture. Providing the assumed correct personality can improve the simulation accuracy and lower the distribution divergence.}
\label{tbl:full_persona_chara}
\end{table*}

\begin{table*}[]
\centering
\resizebox{\linewidth}{!}{%
\begin{tabular}{l|l|rrrrrrrrrrrrrrrrrrrr|r}
\toprule
\textbf{Model} &
  \textbf{Personality} &
  \multicolumn{1}{l}{\textbf{$C_0$}} &
  \multicolumn{1}{l}{\textbf{$C_1$}} &
  \multicolumn{1}{l}{\textbf{$C_2$}} &
  \multicolumn{1}{l}{\textbf{$C_3$}} &
  \multicolumn{1}{l}{\textbf{$C_4$}} &
  \multicolumn{1}{l}{\textbf{$C_5$}} &
  \multicolumn{1}{l}{\textbf{$C_6$}} &
  \multicolumn{1}{l}{\textbf{$C_7$}} &
  \multicolumn{1}{l}{\textbf{$C_8$}} &
  \multicolumn{1}{l}{\textbf{$C_9$}} &
  \multicolumn{1}{l}{\textbf{$C_{10}$}} &
  \multicolumn{1}{l}{\textbf{$C_{11}$}} &
  \multicolumn{1}{l}{\textbf{$C_{12}$}} &
  \multicolumn{1}{l}{\textbf{$C_{13}$}} &
  \multicolumn{1}{l}{\textbf{$C_{14}$}} &
  \multicolumn{1}{l}{\textbf{$C_{15}$}} &
  \multicolumn{1}{l}{\textbf{$C_{16}$}} &
  \multicolumn{1}{l}{\textbf{$C_{17}$}} &
  \multicolumn{1}{l}{\textbf{$C_{18}$}} &
  \multicolumn{1}{l|}{\textbf{$C_{19}$}} &
  \multicolumn{1}{l}{\textbf{Avg.}} \\ \midrule
\multirow{2}{*}{\textbf{GLM-4-air}} &
  \textbf{role} &
  60 &
  65 &
  70 &
  75 &
  75 &
  80 &
  50 &
  70 &
  75 &
  75 &
  70 &
  65 &
  65 &
  75 &
  90 &
  75 &
  65 &
  60 &
  75 &
  65 &
  70 \\
 &
  \textbf{type} &
  15 &
  15 &
  15 &
  25 &
  5 &
  30 &
  10 &
  5 &
  25 &
  15 &
  15 &
  30 &
  5 &
  20 &
  20 &
  15 &
  10 &
  5 &
  10 &
  5 &
  15 \\ \midrule
\multirow{2}{*}{\textbf{Doubao-1.5-pro}} &
  \textbf{role} &
  0 &
  90 &
  85 &
  100 &
  0 &
  55 &
  70 &
  0 &
  100 &
  100 &
  0 &
  100 &
  100 &
  100 &
  100 &
  100 &
  100 &
  5 &
  100 &
  100 &
  70 \\
 &
  \textbf{type} &
  0 &
  90 &
  0 &
  75 &
  0 &
  50 &
  0 &
  0 &
  95 &
  0 &
  0 &
  5 &
  100 &
  10 &
  100 &
  0 &
  0 &
  0 &
  100 &
  0 &
  31 \\ \bottomrule
\end{tabular}%
}
\caption{The detailed personality prediction accuracy (\%) result for the U.S. culture simulation. The model can predict more accurately in terms of role level but has difficulty predicting the smaller granular personality types.}
\label{tbl:persona_acc}
\end{table*}

\end{document}